\pdfoutput=1

\documentclass[11pt]{article}

\usepackage{EMNLP2022}

\usepackage{times}
\usepackage{latexsym}
\usepackage{graphicx}
\usepackage{bm}
\usepackage{amsmath}
\usepackage{amssymb}
\usepackage{multirow}
\usepackage{colortbl}
\usepackage{graphicx}
\usepackage{booktabs}
\usepackage{longtable}
\usepackage[T1]{fontenc}

\usepackage[utf8]{inputenc}

\usepackage{microtype}

\usepackage{inconsolata}
\title{Cross-Linguistic Syntactic Difference in Multilingual BERT:\\
How Good is It and How Does It Affect Transfer?}

\author{Ningyu Xu$^{1, 2}$, Tao Gui$^{2}$\thanks{\ \ Corresponding author.} , Ruotian Ma$^{1}$, Qi Zhang$^{1}$, Jingting Ye$^{3,4}$, Menghan Zhang$^{2}$, Xuanjing Huang$^{1,2}$\\
$^{1}$School of Computer Science, Fudan University\\
$^{2}$Institute of Modern Languages and Linguistics, Fudan University\\
$^{3}$Department of Chinese Language and Literature, Fudan University\\
$^{4}$Department of Linguistic and Cultural Evolution, Max Planck Institute for Evolutionary Anthropology\\
  \texttt{nyxu22@m.fudan.edu.cn} \hspace{0.5cm}
  \texttt{\{tgui,rtma19,qz,yejingting,mhzhang,xjhuang\}@fudan.edu.cn}
}

\begin{document}
\maketitle
\begin{abstract}

Multilingual BERT (mBERT) has demonstrated considerable cross-lingual syntactic ability, whereby it enables effective zero-shot cross-lingual transfer of syntactic knowledge. The transfer is more successful between some languages, but it is not well understood what leads to this variation and whether it fairly reflects difference between languages. In this work, we investigate the distributions of grammatical relations induced from mBERT in the context of 24 typologically different languages. We demonstrate that the distance between the distributions of different languages is highly consistent with the syntactic difference in terms of linguistic formalisms. Such difference learnt via self-supervision plays a crucial role in the zero-shot transfer performance and can be predicted by variation in morphosyntactic properties between languages. These results suggest that mBERT properly encodes languages in a way consistent with linguistic diversity and provide insights into the mechanism of cross-lingual transfer.

\end{abstract}

\section{Introduction}

Cross-lingual transfer aims to address the huge linguistic disparity in NLP by transferring the knowledge acquired in high-resource languages to low-resource ones, where pretrained multilingual encoders, such as Multilingual BERT (mBERT) \cite{devlin_bert_2019}, have proven a powerful facilitator. Compared to other approaches learning certain cross-lingual alignment in a supervised \cite{10.5555/3045118.3045199, mikolov2013exploiting, faruqui-dyer-2014-improving} or unsupervised \cite{artetxe-etal-2017-learning,zhang-etal-2017-earth,lample_word_2018} manner, mBERT directly learns to encode different languages in a shared representation space through self-supervised joint training, dispensing with explicit alignment. It has exhibited notable cross-lingual ability and can perform effective zero-shot cross-lingual transfer across a variety of downstream tasks, albeit the performance varies \cite{wu_beto_2019, pires_how_2019}. 

The simplicity and efficacy of mBERT are crucial for cross-lingual transfer and have sparked interest in investigating the reason for its success. Previous work has looked into its representation space and found that mBERT automatically performs certain alignment across languages \cite{cao_multilingual_2019, gonen_its_2020, conneau_emerging_2020, chi_finding_2020}. The extent of alignment is shown correlated with the transfer performance \cite{muller_first_2021}. Despite these insights into the source of the transfer, it is also intriguing why different languages are aligned to varying degrees and what implication such variation bears. Another line of work has demonstrated that the zero-shot transfer performance is affected by certain linguistic features such as word order \cite{pires_how_2019, karthikeyan2019cross}, whereas the underlying mechanism is left unexplored. Taken together, it remains unclear how different aspects of cross-linguistic differences impact the representations and further affect the cross-lingual transfer of different tasks.

In this paper, we focus on the syntactic level and investigate the cross-lingual transfer of mBERT based on 24 typologically distinct languages, with the purpose of figuring out the following questions:

\textbf{Q1}: \textbf{Does mBERT properly induce cross-linguistic syntactic difference via self-supervision?} The distance between distributions over mBERT representations of grammatical relations in different languages can be used to evaluate the syntactic difference between languages encoded in mBERT (Section~\ref{sec:syntactic-difference}). We compare it with the cross-linguistic syntactic difference in terms of linguistic formalisms for validation and rely on it to investigate the cross-lingual ability of mBERT.

\textbf{Q2}: \textbf{How does the syntactic difference learnt by mBERT impact its cross-lingual transfer?} The zero-shot cross-lingual transfer is typically realized through fine-tuning the pretrained multilingual model on a certain source language. We analyze the change pretraining and fine-tuning brought to the distance between distributions (i.e., the syntactic difference between languages in mBERT) to understand the mechanism behind the transfer (Section~\ref{sec: mechanism}). 

\textbf{Q3}: \textbf{If and to what extent do various morphosyntactic properties impact the transfer performance?} We then investigate the reason for the variation in the transfer performance based on syntactic-related linguistic properties. We exploit all the morphosyntactic properties available in the World Atlas of Language Structures (WALS) \citep{wals} and examine the extent to which variation in them impacts the distance between distributions and further affects the transfer performance through regression analysis (Section~\ref{sec: typology}). 

Our quantitative results and qualitative analysis demonstrate that: 

1) The distance between distributions of grammatical relations in mBERT is highly consistent with the cross-linguistic syntactic difference in the context of linguistic formalisms. 2) The syntactic difference learnt during pretraining plays a crucial role in the zero-shot cross-lingual transfer of dependency parsing. While fine-tuning on a specific language augments the transfer with task-specific knowledge, it can distort the established cross-linguistic knowledge. 3) Variation in morphosyntactic properties is predictive of the syntactic difference in mBERT, which further impacts the transfer performance. Encouragingly, these linguistic features can be exploited to optimize the cross-lingual transfer, whereby we can efficiently select the best language for fine-tuning without the need for any dataset.\footnote{Our code is available at \url{https://github.com/ningyuxu/cl-syntactic-difference-mbert}.}

\section{A Measure of Cross-Linguistic Syntactic Difference in mBERT}
\label{sec:syntactic-difference}

mBERT learns to encode different languages in a shared representation space, which provides a basis for cross-linguistic comparison. However, the syntactic properties of a language are not explicitly realized at a word or sentence level. To bridge the gap between the linguistic knowledge at a language level and the word-level contextual representations, we look into the distributions over mBERT representations of different languages. We first derive representations of syntactic information (i.e., grammatical relations) from mBERT and then use the divergence between distributions over the representations to measure the language-wide difference encoded in it. Finally, we compare the measured difference with the cross-linguistic syntactic difference in terms of formal syntax to examine whether mBERT properly induces syntactic difference via self-supervision.

\subsection{Method}
\label{sec:measure-syntactic-difference-method}

\paragraph{Multilingual BERT}

mBERT is a Transformer-based \cite{vaswani2017attention} neural language model, which has the same architecture as BERT-Base but is pretrained on a concatenation of monolingual Wikipedia corpora from 104 languages. For each input sentence tokenized into a sequence of $n$ tokens $w_{1:n}$, mBERT runs them through an embedding layer and 12 layers of transformer encoders, producing a sequence of contextual representations $\mathbf{h}_{1:n}^{\ell}$ for each token at each layer $\ell$, where $1 \le \ell \le 12$. As there is no explicit cross-lingual alignment provided during the entire training procedure, it is intriguing \textbf{how common linguistic properties vary across languages in mBERT representation space}.

\paragraph{Representations of grammatical relations in mBERT}

We adopt the framework of Universal Dependencies (UD) \cite{de_marneffe_universal_2021} in describing abstract syntactic structure across typologically diverse languages, where the dependency grammatical relations are universal and allow for cross-linguistic comparison. In the light of work of the structural probe \cite{hewitt_structural_2019, chi_finding_2020}, we use the difference between mBERT representations of a head-dependent pair of words $( w_{\textrm{\footnotesize head}}, w_{\textrm{\footnotesize dep}})$ to represent the grammatical relation between them: 
\begin{equation}
    \mathbf{d}_{(\textrm{\footnotesize head}, \textrm{\footnotesize dep})}^{\ell} = \mathbf{h}_{\textrm{\footnotesize head}}^{\ell} - \mathbf{h}_{\textrm{\footnotesize dep}}^{\ell}, \label{formula:head-dep-representation}
\end{equation}
and verify its effectiveness through a linear classifier decoding the grammatical relation from it. We then visualize the representations $\mathbf{d}_{(\textrm{\footnotesize head}, \textrm{\footnotesize dep})}^{\ell}$ to get a qualitative understanding of the grammatical information encoded in them\footnote{See Appendix~\ref{sec:grammatical-relation-probe} for details.}.

\paragraph{Evaluation of cross-linguistic syntactic difference in mBERT}
\label{sec:syntactic-difference-mBERT}

To evaluate the language-wide difference in terms of grammatical relations, we abstract away from single sentences and look into the distributions of representations. Formally, we regard the dataset in language $L$ as a set of $N$ feature-label pairs, i.e., $\mathcal{D}_{L} = \left \{\left(\mathbf{x}^{(i)}, \mathbf{y}^{(i)}\right ) \right \}_{i=1}^{N} \sim P_{L} \left(\mathbf{x}, \mathbf{y}\right)$, where feature $\mathbf{x}$ is our representation $\mathbf{d}_{(\textrm{head}, \textrm{dep})}$ and the label $\mathbf{y}$ is the gold grammatical relation between the word pair $(w_{\textrm{head}}, w_{\textrm{dep}})$. $P_{L} \left(\mathbf{x}, \mathbf{y}\right)$ denotes the joint distribution over the feature-label space. We define the syntactic difference between $L_A$ and $L_B$ ($d_{\mathcal{S}}(L_A, L_B)$) as the distance between their joint distributions: 
\begin{equation}
    d_{\mathcal{S}}(L_A, L_B) \triangleq d\left(P_{L_A}\left(\mathbf{x}, \mathbf{y}\right), P_{L_B}\left(\mathbf{x}, \mathbf{y}\right)\right). \label{formula:syntactic-diffrence-mbert}
\end{equation}
The optimal transport dataset distance (OTDD) \cite{alvarez-melis_geometric_2020} is employed for the estimation of the distance\footnote{See Appendix~\ref{sec:appendix-otdd} for details.}, as it has a solid theoretical footing, discards extra parameters, and yields distance both between datasets and between labels, benefiting fine-grained analysis of the representation space.

\paragraph{Validation of cross-linguistic syntactic difference in mBERT}

We validate the effectiveness of our measure through comparison with the cross-linguistic syntactic difference in the context of linguistic formalisms. We adopt the formal syntactic distance provided in \citet{ceolin_formal_2020}, which is measured based on the theory of Principles-and-Parameters developed since \citet{Chomsky+2010}. It compares the syntactic structure of different languages through a finite set of universal abstract grammatical parameters characterizing possible cross-linguistic differences, which in principle enables a systematic comparison between syntax of different languages \cite{longobardi_evidence_2009}. In detail, each parameter is coded as a binary value, and a language $L$ is represented by the list of parameters $S_{L}$ it takes. The formal syntactic distance between language $L_{A}$ and $L_{B}$ is measured by Jaccard distance \cite{jaccard-1901} between them: 
\begin{equation}
    d_{\mathcal{F}}(L_{A}, L_{B}) \triangleq d_{\mathrm{Jaccard}}(S_{L_{A}}, S_{L_{B}}). \label{formula:formal-distance}
\end{equation}

\subsection{Experimental Setup}

\paragraph{Data}

The data for all our experiments is from UD treebanks. We adopt all the grammatical relations defined in it except for \emph{root} as it does not denote relations between words. We select 24 typologically different languages covering a reasonable variety of language families, which are Arabic, Bulgarian, German, Greek, English, Spanish, Estonian, Persian, Finnish, French, Hebrew, Hindi, Italian, Japanese, Korean, Latvian, Dutch, Polish, Portuguese, Romanian, Russian, Turkish, Vietnamese and Chinese\footnote{Constrained by the availability of UD datasets and mBERT's pretraining, many languages belong to the Indo-European family. See Appendix~\ref{sec:data} for the datasets we use.}. 


\paragraph{Baseline}
\label{sec: baseline}

We compare mBERT with the following two baselines:
\begin{itemize}
  \item \textbf{\textsc{mBERT0}} The layer 0 of mBERT, which does not involve any contextual information.
  \item \textbf{\textsc{mBERTrand}} A model same as mBERT but without pretrained weights. The subword embeddings remain unchanged.
\end{itemize}

\subsection{Results}

\paragraph{Evaluating cross-linguistic syntactic difference in mBERT}

\begin{figure}[t]
\centering
  \includegraphics[width=7.5cm]{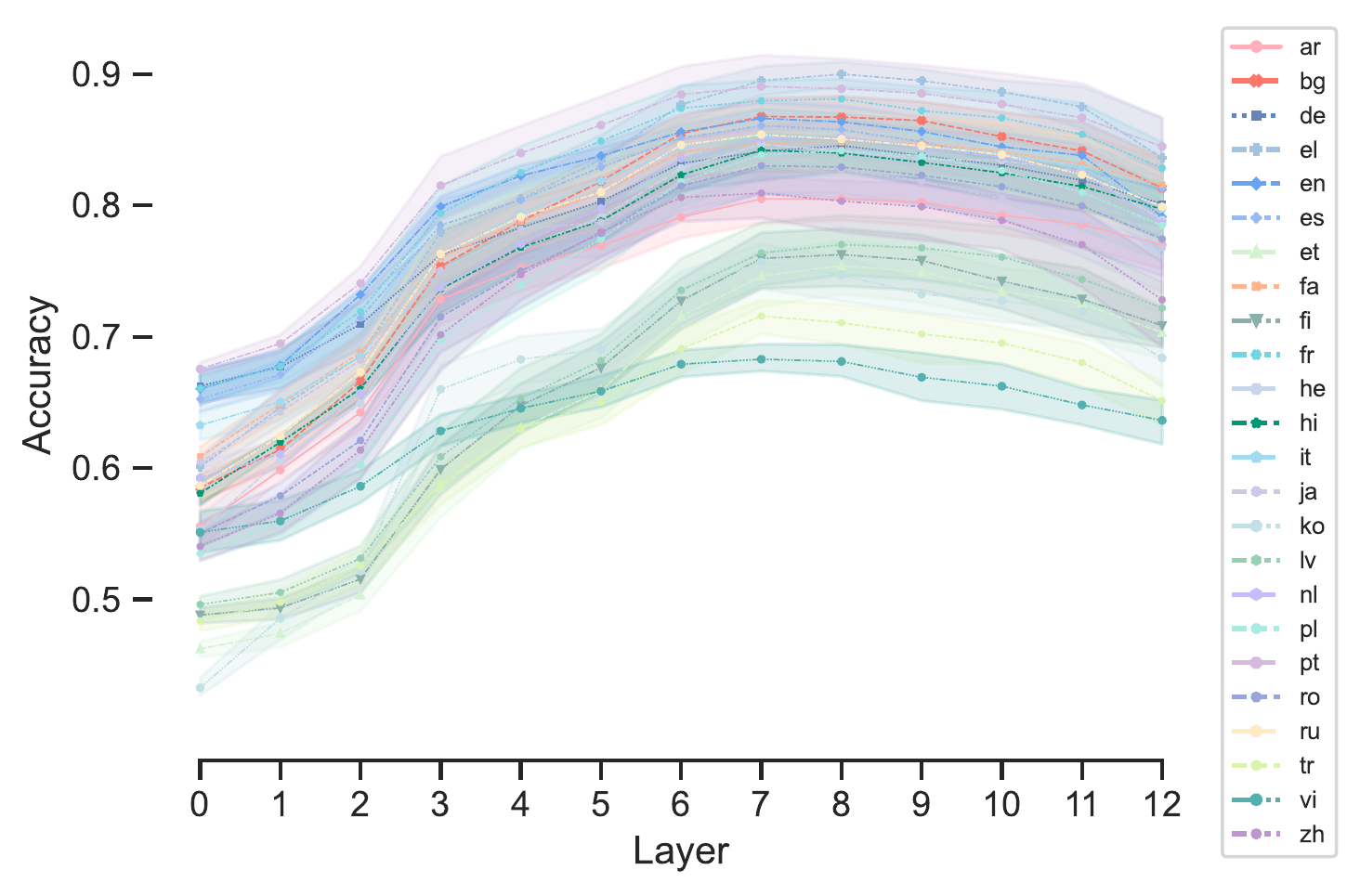}
  \caption{Accuracy in recovering grammatical relations of different languages across the layers of mBERT. The colored bands denote $95\%$ confidence intervals.} \label{fig:probe-acc}
\end{figure}


The probing result (Figure~\ref{fig:probe-acc}) demonstrates that \textbf{grammatical relation can be successfully extracted from the representations} computed based on our method in contrast to baselines\footnote{See Appendix~\ref{sec:grammatical-relation-probe} Table~\ref{tab:probe-comparison-baseline} for comparison with baselines.}. The 7th and 8th layer are most effective in encoding grammatical relations across the languages.



\begin{figure}[t]
\centering
  \includegraphics[width=7.5cm]{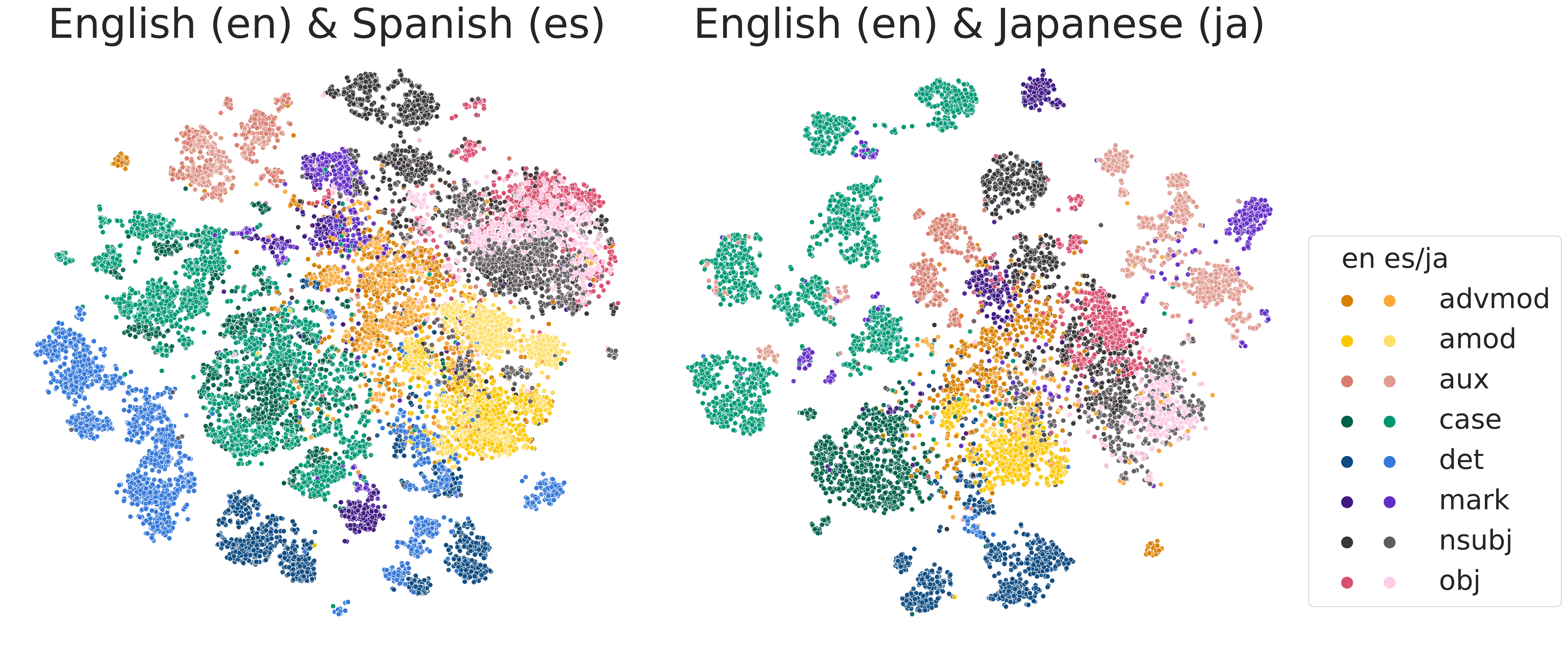}
  \caption{Visualization of the representations of different grammatical relations derived from the 7th layer of mBERT. English is shown more similar to Spanish than to Japanese as to the distributions of grammatical relations such as \emph{case}, \emph{obj} and \emph{aux}.} \label{fig:visual-en-es-ja}
\end{figure}

The representations we derive from the 7th layer of mBERT generally form clusters reflecting their grammatical relations (Figure~\ref{fig:visual-en-es-ja}). Moreover, we can find that the distributions of different languages differ and such difference reflects certain difference between languages. The representations of the same grammatical relations better clustered together between English and Spanish than between English and Japanese, in line with the fact that English is more similar to Spanish than to Japanese at the syntactic level. 


\paragraph{Validating cross-linguistic syntactic difference in mBERT}
\label{sec:syntactic-difference-validation}

\begin{figure}[t]
\centering
  \includegraphics[width=7.5cm]{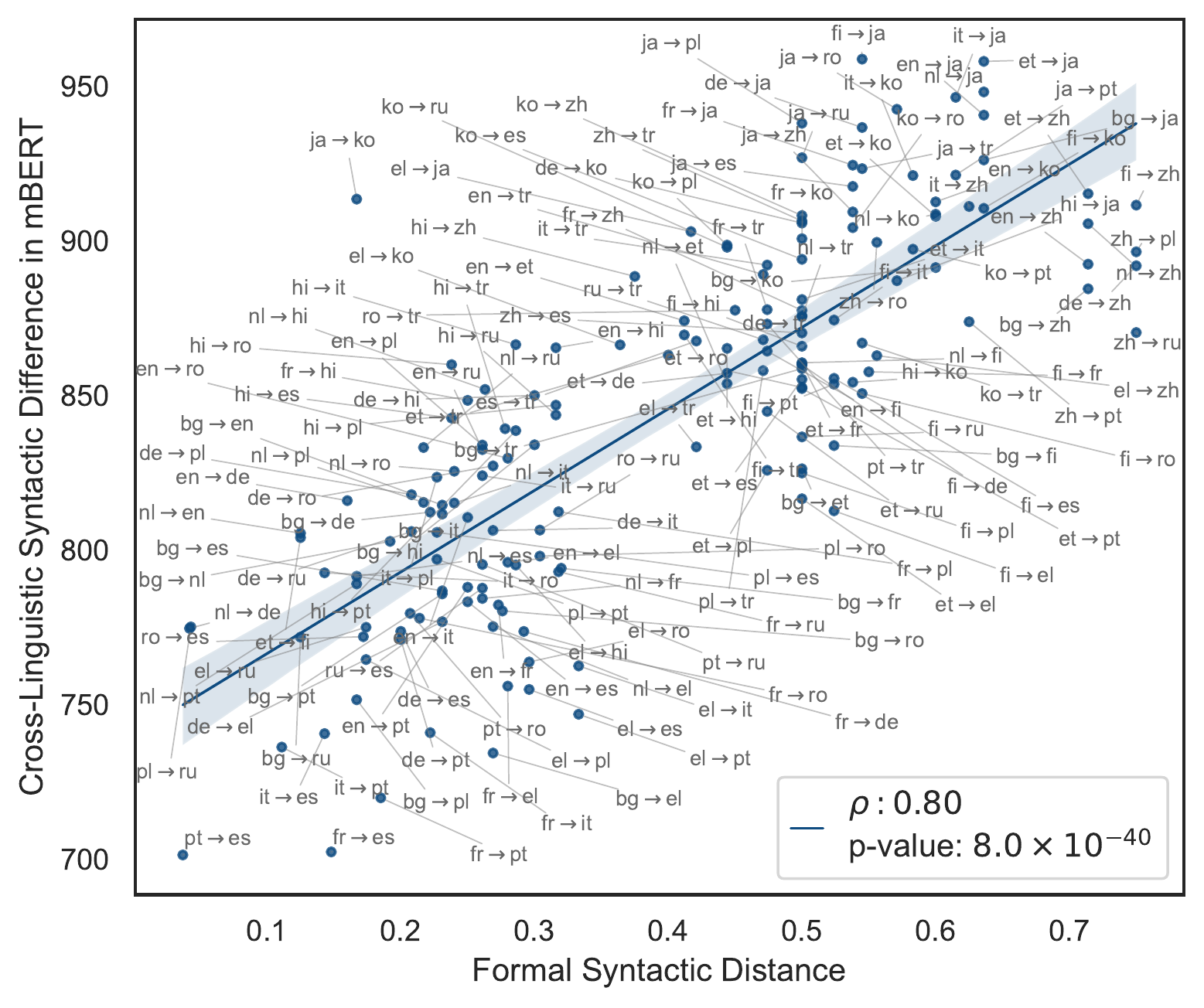}
  \caption{Comparison of the formal syntactic distance and the cross-linguistic syntactic difference induced from mBERT, evaluated through Spearman’s correlation.} \label{fig:formal-dist-otdd}
\end{figure}


The cross-linguistic syntactic difference measured based on mBERT shows significantly high correlation with the formal syntactic distance (Figure~\ref{fig:formal-dist-otdd}). And the correlation is higher in the 7th layer ($\rho = 0.80$) than baselines ($\rho = 0.72$ for \textsc{mBERT0} and $0.68$ for \textsc{mBERTrand}), which indicates that \textbf{mBERT properly induces difference in syntactic structure via self-supervision}.

\subsection{Discussion}

\begin{figure}[t]
\centering
  \includegraphics[width=7.5cm]{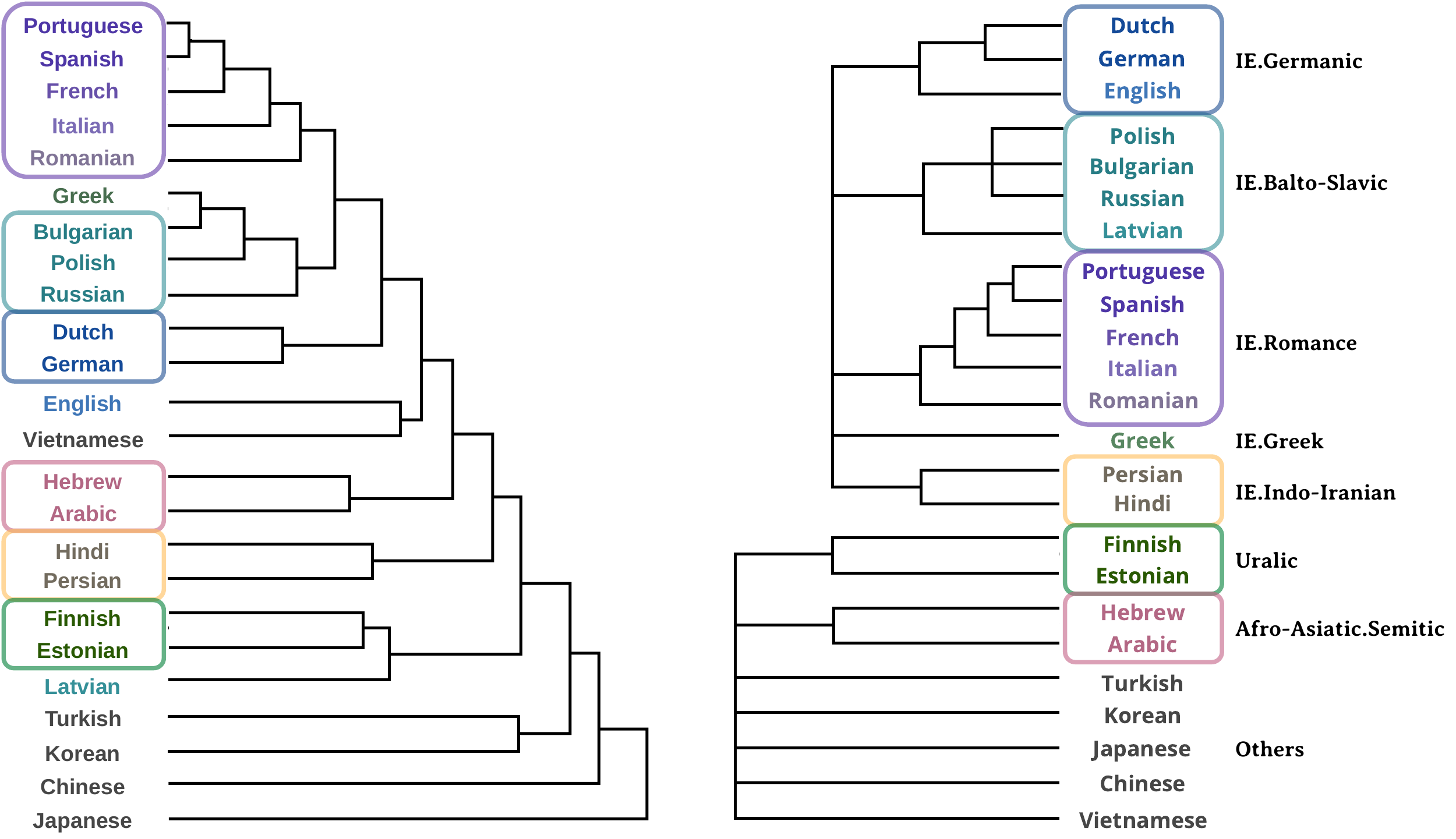}
  \caption{Left: Hierarchical clustering based on cross-linguistic syntactic difference derived from mBERT. Right: The gold phylogenetic tree from Glottolog \cite{harald_hammarstrom_2021_5772642}. IE stands for the Indo-European family.} \label{fig:cluster-otdd}
\end{figure}

\textbf{Grammatical relations can be largely derived from the representations computed based on our method, but to different degrees.} As shown in the probing result (Figure~\ref{fig:probe-acc}), the representations are less effective in encoding syntactic knowledge for languages such as Turkish, Hebrew, Estonian, Finnish, Korean and Chinese, where the former four have rich morphology and the latter are tokenized with CJK characters\footnote{Additionally, the deficiency in Vietnamese may result from lack of training data as its treebank is relatively small.}. Previous work has demonstrated similar disparity in mBERT \cite{chi_finding_2020,mueller_cross-linguistic_2020} and suggests that the inadequacy in tokenization can be a possible reason \cite{rust_how_2021}.

\textbf{While the syntactic difference induced from mBERT is highly consistent with the distance in formal grammar, certain deviation can be observed,} especially for languages poorly represented where the probe classifier achieves a relatively lower performance.

We further perform a hierarchical clustering based on our measure to understand the relationship between languages it reveals. \textbf{Languages in the same family are generally clustered together, analogous to conventional understanding in linguistic taxonomy, while there exist discrepancies} regarding languages such as Vietnamese and Latvian (Figure~\ref{fig:cluster-otdd}). Besides the deficiency in representations, they might stem from i) the sampling bias in the UD treebanks, especially for low-resource languages such as Vietnamese, and ii) the difference between languages in terms of dependency grammar better reflecting grammatical diversity. For instance, though belonging to the Indo-European family, Latvian bears structural similarities to Finno-Ugric languages \cite{kalnaca2014typological}. Such result is in line with previous work showing certain correlation between grammatical typology and historical relatedness \cite{doi:10.1126/science.1114615,Wichmann2007HowTU,longobardi_evidence_2009,abramov-dependency-2011} and suggests that the relationship between languages in terms of syntax should be properly reflected in mBERT representation space.

\section{Mechanism behind Cross-Lingual Transfer}
\label{sec: mechanism}

The training procedure of zero-shot transfer typically involves two steps: pretraining on a multilingual corpus and fine-tuning on a specific source language. To understand the mechanism behind the zero-shot transfer and why the transfer performance varies across languages, we look into the change they bring to the syntactic difference in mBERT. Specifically, we first compare the syntactic difference learnt during pretraining with the transfer performance to evaluate the impact of pretraining on the transfer and then examine how fine-tuning on a specific language changes the syntactic difference.

\subsection{Method}

\paragraph{Analyzing the effect of pretraining}

We investigate what effect the syntactic difference learnt during pretraining has on the transfer performance through a correlation analysis. The performance of dependency parsing is measured by labeled attachment score (LAS). Let 
\begin{equation}
    drop(L_{S}, L_{T}) \triangleq \textrm{LAS}_{L_S} - \textrm{LAS}_{L_T} \label{eqn:transfer-performance}
\end{equation}
denote the drop in LAS when transferring the model fine-tuned on a source language $L_S$ to a target language $L_T$, we compare it with the syntactic difference $d_{\mathcal{S}}^{(\textrm{\footnotesize pre})}(L_{S}, L_{T})$ measured based on (\ref{formula:syntactic-diffrence-mbert}) in pretrained mBERT.

\paragraph{Analyzing the effect of fine-tuning}

To understand how fine-tuning on a source language impacts the zero-shot transfer of the dependency parsing task, we investigate the change it brings to the syntactic difference between the source and target languages in mBERT. We first visualize mBERT representations of grammatical relations\footnote{We use the same method of visualization as in Section~\ref{sec:measure-syntactic-difference-method}. See Appendix~\ref{sec:grammatical-relation-probe} for details.} before and after fine-tuning to get a qualitative understanding, and then quantitatively compare the syntactic difference in pretrained mBERT and mBERT fine-tuned on the source language. 

To further explore whether the change in the syntactic difference impacts the variation in transfer performance among different target languages, we perform a correlation analysis of their syntactic difference with the source language before and after fine-tuning. We also compare the change that fine-tuning on different source languages brings to the syntactic difference to understand how fine-tuning on a particular language may affect the overall cross-linguistic syntactic knowledge in mBERT.

\subsection{Experimental Setup}

Following the setup of \citet{wu_beto_2019}, we use the parser with deep biaffine attention mechanism \cite{dozat_deep_2017} as the task-specific layer on top of mBERT for dependency parsing, which has been shown to perform the best on average across typologically different languages \citep{ahmad_difficulties_2019}. Instead of providing gold Part-of-Speech (POS) tags, we train a linear model to predict POS tags on the source side, and apply it to the target language. We employ this strategy to avoid introducing additional cross-lingual information, as we focus on the cross-lingual ability of mBERT itself.

We take the 7th layer of pretrained mBERT as information about grammatical relations is best manifested here. For fine-tuned models, we focus on the 12th layer as the representations here are directly fed to the parser and impact the transfer performance.

\subsection{Results}
\label{sec:mechanism-result}

\begin{figure}[t]
\centering
  \includegraphics[width=7.5cm]{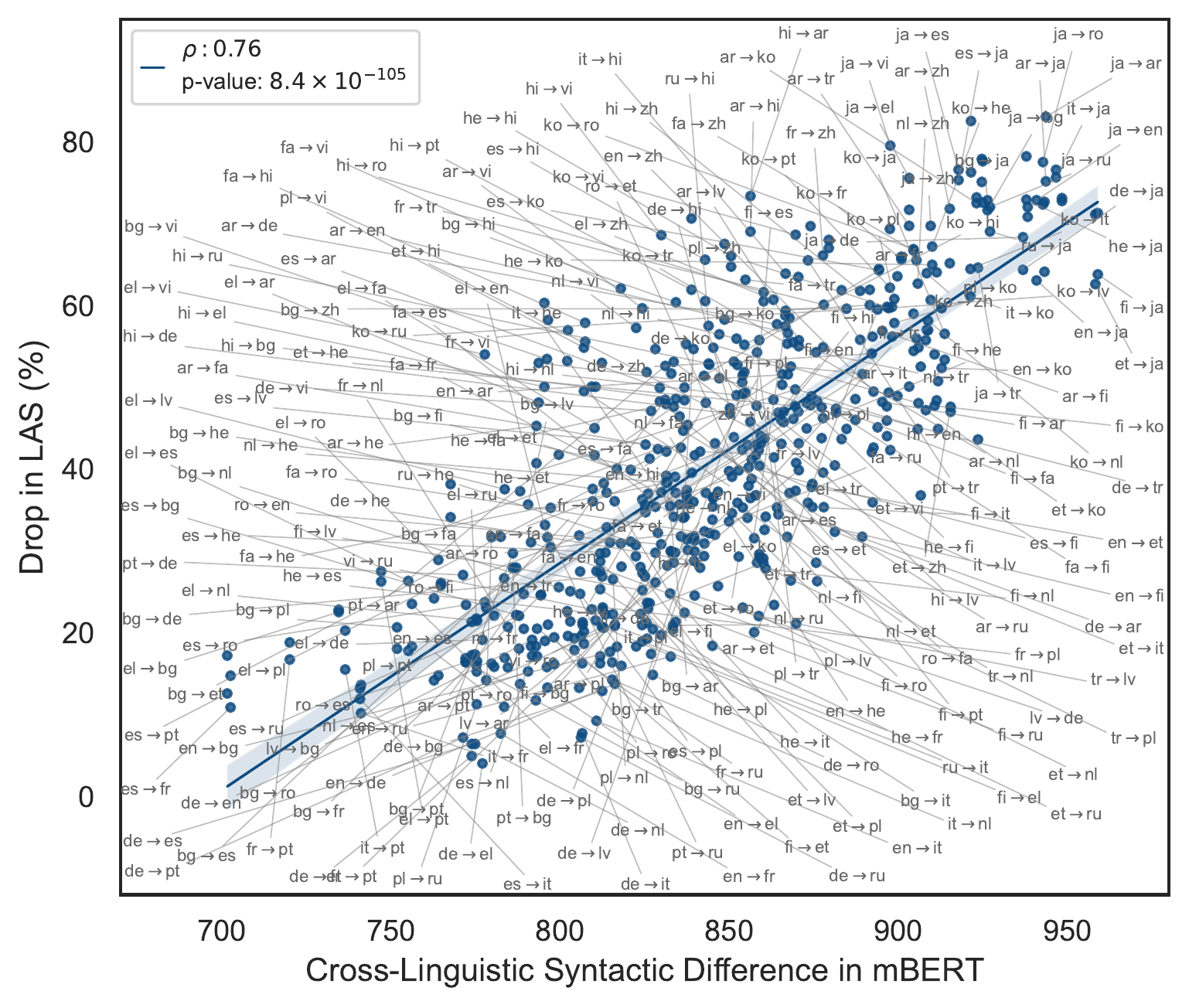}
  \caption{Comparison of the cross-linguistic syntactic difference in pretrained mBERT and drop in zero-shot cross-lingual transfer performance (LAS).} \label{fig:otdd-pretrain-vs-las}
\end{figure}

\paragraph{Effect of pretraining}

The syntactic difference acquired during pretraining strongly correlates with the drop in LAS across typologically diverse languages (Figure~\ref{fig:otdd-pretrain-vs-las}), in contrast to the baselines ($\rho = 0.51$ for \textsc{mBERT0} and $0.45$ for \textsc{mBERTrand}). The result suggests that, with a given source language, \textbf{the syntactic difference learnt during pretraining plays a crucial role in the cross-lingual transfer performance}.

\paragraph{Effect of fine-tuning}

\begin{figure}[t]
\centering
  \includegraphics[width=7.5cm]{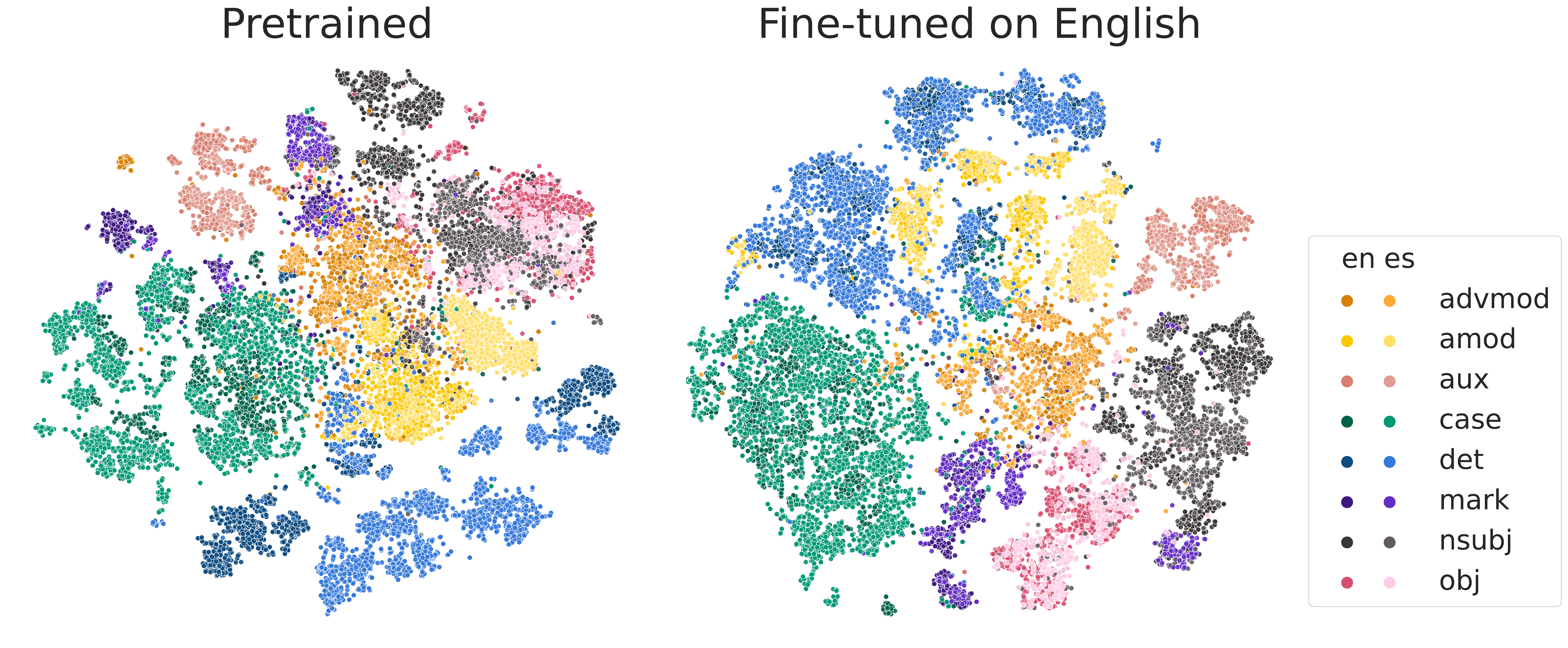}
  \caption{Visualization of the representations of grammatical relations in English (en) and Spanish (es) derived from pretrained mBERT and mBERT fine-tuned on English. Representations of the same grammatical relation in different languages better cluster together after fine-tuning.} \label{fig:visualize-en-es-pre-finetune}
\end{figure}

\begin{figure}[t]
\centering
  \includegraphics[width=7.5cm]{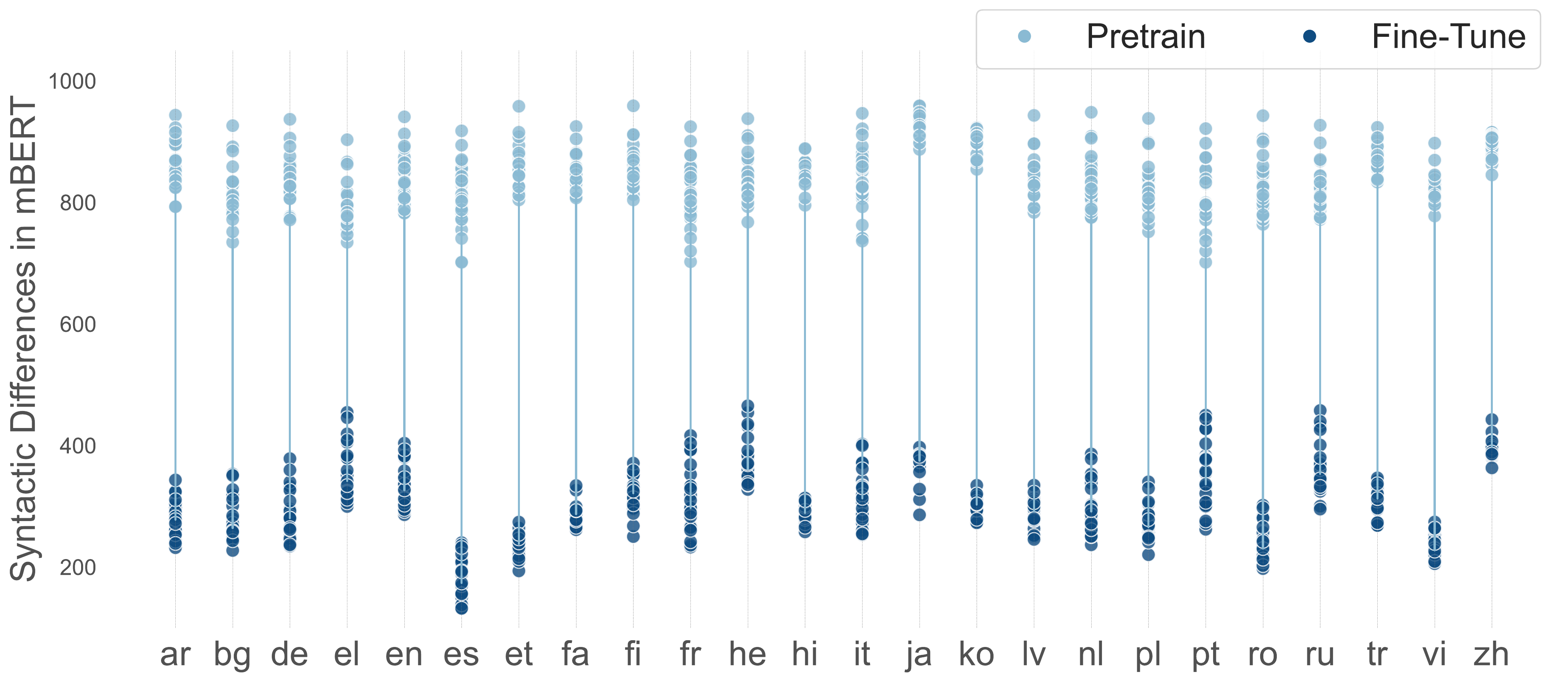}
  \caption{The syntactic difference in mBERT before and after fine-tuning on the language on the x-axis. Each point represents the syntactic difference between the source language on the x-axis and another language.} \label{fig:otdd-shift}
\end{figure}

\begin{figure}[t]
\centering
  \includegraphics[width=7.5cm]{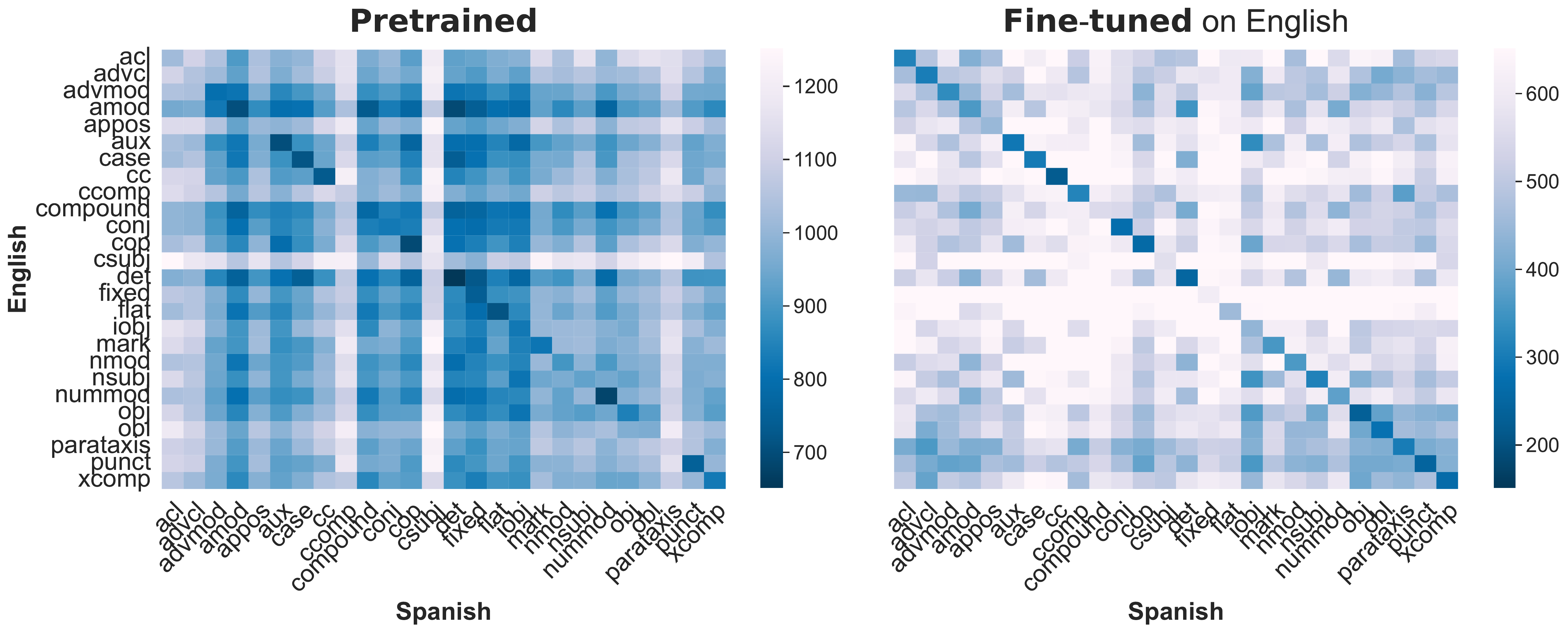}
  \caption{Distance between distributions of grammatical relations in English and Spanish before and after fine-tuning on English. The distance between the same grammatical relation becomes much smaller after fine-tuning, indicating a task-specific improvement in transfer.} \label{fig:dependency-dist-en-es}
\end{figure}

After fine-tuning on English, representations of the same grammatical relation better cluster together (Figure~\ref{fig:visualize-en-es-pre-finetune}), indicating a task-specific improvement in both the source and target languages. Our quantitative analysis reveals that the syntactic difference with the source language in mBERT generally decreases after fine-tuning (Figure~\ref{fig:otdd-shift}), where the distance between the same grammatical relations decrease much more drastically than the others (Figure~\ref{fig:dependency-dist-en-es}). These results, together, suggest that fine-tuning facilitates the zero-shot cross-lingual transfer with task-specific knowledge.

\begin{figure}[t]
\centering
  \includegraphics[width=7.5cm]{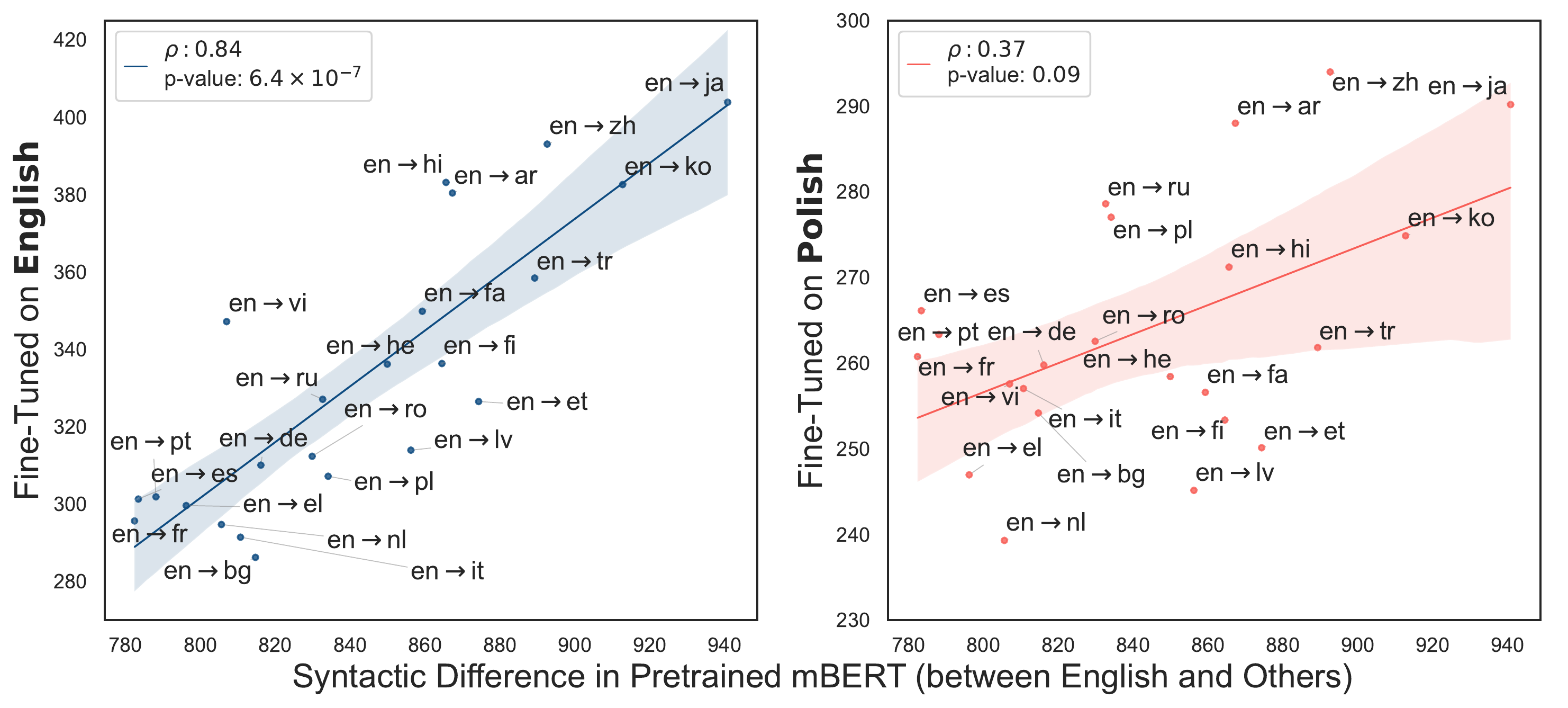}
  \caption{Left: Comparison of the syntactic difference between English and other languages derived from pretrained mBERT (x-axis) and mBERT fine-tuned on English (y-axis). Right: Comparison of the syntactic difference between English and others derived from pretrained mBERT (x-axis) and mBERT fine-tuned on Polish (y-axis). The syntactic difference in mBERT fine-tuned on Polish is not significantly correlated with that in the pretrained model ($p > 0.05$).} \label{fig:en-vs-pl}
\end{figure}

Through a correlation analysis, we find an approximately linear relationship between the syntactic difference with the source language before and after fine-tuning. Namely, \textbf{fine-tuning on a specific language benefits other languages according to the similarity between them learnt during pretraining}. However, \textbf{it can distort the overall cross-linguistic syntactic knowledge}, especially for languages with a bigger difference. Figure~\ref{fig:en-vs-pl} shows that the syntactic difference with English is worse correlated with $d_{\mathcal{S}}^{(\textrm{\footnotesize pre})}(\textrm{en}, \cdot)$ when fine-tuning on typologically distant languages such as Polish than on English, indicating that \textbf{the relationship among languages can be deformed when augmenting the pretrained model with syntactic knowledge in a single language}.

\subsection{Discussion}

Our experiment results are complementary to previous work in the monolingual setting, which has shown that fine-tuning benefits downstream tasks with clearer distinction between samples belonging to different labels but also largely preserves the original spatial structure of the pretrained model \cite{merchant-etal-2020-happens,zhou-srikumar-2022-closer}. In mBERT, we can further explore how fine-tuning on a specific language impacts the representations of other languages, i.e., samples in different domains. While for that language, fine-tuning augments the effect of pretraining and benefits the transfer, it distorts the established cross-linguistic knowledge especially for languages with a larger divergence in distributions.

\section{Impact of Linguistic Diversity}
\label{sec: typology}

To better understand the cross-linguistic syntactic difference learnt by mBERT, we employ the structural and functional features in linguistic typology which allows for description of linguistic diversity and analyze to what extent variation in these features affects the syntactic difference in mBERT. We further examine whether these features can be exploited to select better source languages and thus benefit cross-lingual transfer.

\subsection{Method}

\paragraph{Typological features}

We exploit all the morphosyntactic features available in WALS \cite{wals}, covering areas including Morphology, Nominal Categories, Verbal Categories, Nominal Syntax, Word Order, Simple Clauses, and Complex Sentences.\footnote{We filter out the features which have missing values for all the languages we study, which results in a total of 116 features. See Appendix~\ref{sec:appendix-typo} for all the features we use.}


\paragraph{Evaluation of difference in typological features}

For each feature $f$, there are between 2 to 28 different values in WALS and they may not be mutually exclusive. We regard each feature as a vector $\mathbf{v}_{f}^{L} = [v_{1}^{L},\cdots,v_{m}^{L}]$ where $m$ is the number of possible values for a feature $f$ and each entry $v_{i}^{L} (i=1, \cdots, m)$ typically represents a binary value that a language $L$ may take (see Table~\ref{tab:wals-feature-example} for an example). We use cosine distance to measure the difference between language $L_{A}$ and $L_{B}$ in this feature: 
\begin{equation}
    d_{f}(L_{A},L_{B}) \triangleq 1 - \cos \left(\mathbf{v}_{f}^{L_{A}}, \mathbf{v}_{f}^{L_{B}} \right).
\end{equation}
The overall difference between $L_{A}$ and $L_{B}$ is represented by 
\begin{equation}
    \mathbf{d}_{F}(L_{A},L_{B}) = [d_{f_{1}}, \cdots, d_{f_{n}}],
\end{equation}
where $n = 116$ is the total number of features.

\begin{table}
\centering
\begin{tabular}{cccc}
\hline
\textbf{Language} & \textbf{NRel} & \textbf{RelN} & \textbf{Correlative}\\
\hline
English   & 1 & 0 & 0 \\
Hindi     & 0 & 0 & 1 \\
Hungarian & 1 & 1 & 0 \\
Japanese  & 0 & 1 & 0 \\
\hline
\end{tabular}
\caption{A truncated example of WALS feature 90A: Order of Relative Clause (Rel) and Noun (N). Each entry typically takes a binary value for a particular language. For Hungarian, there is not a dominant type of the order of Rel and N, and instead, both NRel and RelN exist.}
\label{tab:wals-feature-example}
\end{table}

\paragraph{Regression analysis}

Given the observable correlation and potential interdependence between these features\footnote{For instance, implicational universals of word order \cite{Greenberg1990SomeUO, Dryer1992TheGW}) may be driven by some universal constraints \cite{levshina_token-based_2019, hahn_universals_2020}.}, we use a gradient boosting regressor\footnote{\url{https://scikit-learn.org/stable/modules/generated/sklearn.ensemble.GradientBoostingRegressor.html}} combined with impurity-based and permutation importance to analyze the impact of different features, as it is robust to multicollinearity, generally achieves high empirical performance, and is relatively interpretable. The regressor $\mathcal{G}$ takes as input $\mathbf{d}_{F}(L_{A},L_{B})$ and the target is to predict the syntactic difference between them, i.e., $d_{\mathcal{S}}^{(\textrm{\footnotesize pre})}(L_A, L_B)$.

\paragraph{Selection of source languages}

To further examine our findings and also improve the cross-lingual transfer, we extend the regressor to predict the syntactic difference between the $J$ languages we study $\{L_{1},L_{2},\cdots,L_{J}\}$ and another language $L_{K}$ and then test whether $L_{S} = \operatorname*{argmin}_{j}\mathcal{G}\left(\mathbf{d}_{F}(L_{j}, L_{K})\right)$ is among the best source languages for zero-shot cross-lingual transfer.

\subsection{Experimental Setup}

\paragraph{Model and evaluation in regression analysis}

We train the gradient boosting regressor with 100 estimators where each has a maximum depth of three. Its performance is evaluated through the average of $R^{2}$ in 10-fold cross-validation. For feature importance, we report both permutation importance with 30 repeats and impurity-based importance.

\paragraph{Evaluation of source language selection}

We test the effectiveness of our regressor in source language selection on five other languages including Czech, Catalan, Hungarian, Tamil and Urdu. Specifically, each of them is taken as a target language and our goal is to choose the best source language among the languages we select. For each target language, we use the regressor to predict the syntactic differences between it and the source languages, and rank them from low to high to get the predicted ranking of source languages. To get the gold ranking for evaluation, we fine-tune an mBERT on each of the source languages, test it on the target language to obtain the LAS, and rank the scores from high to low. Similar to \citet{lin_choosing_2019}, we use the Normalized Discounted Cumulative Gain \cite{jarvelin2002cumulated} at position 3 (NDCG@3)\footnote{\url{https://scikit-learn.org/stable/modules/generated/sklearn.metrics.ndcg_score.html}} as the evaluation metric. It measures the quality of ranking and yields a score between $0$ and $1$, where the gold ranking gets a score of $1$.

\paragraph{Baseline}

We compare the trained regressor with these baselines:

\begin{itemize}
  \item \textbf{\textsc{Ave}} The average distance of all morphosyntactic features.
  \item \textbf{\textsc{URIEL}} The different kinds of distance provided in \citet{littell_uriel_2017}, including syntactic $d_{\textrm{syn}}$\footnote{The syntactic distance here is the cosine distance between feature vectors derived from typological databases including WALS.}, genetic $d_{\textrm{gen}}$, featural $d_{\textrm{fea}}$, geographic $d_{\textrm{geo}}$, phonological $d_{\textrm{pho}}$ and inventory distance $d_{\textrm{inv}}$.
\end{itemize}

\subsection{Results}

\paragraph{Regression Analysis}

The $R^{2}$ score of the regressor reaches $85\%$, showing that \textbf{the differences in morphosyntactic features are predictive of the syntactic difference between languages learnt by mBERT}. Additionally, that the correlation score ($\rho = 0.89$) between the predicted and the computed syntactic difference is higher than baselines ($\rho = 0.58$ for \textsc{Average} and $0.68$ for $d_{\textrm{syn}}$ in \textsc{URIEL}) suggests that these features should be treated with different importance.

\paragraph{Feature importance}

\begin{figure}[t]
\centering
  \includegraphics[width=8cm]{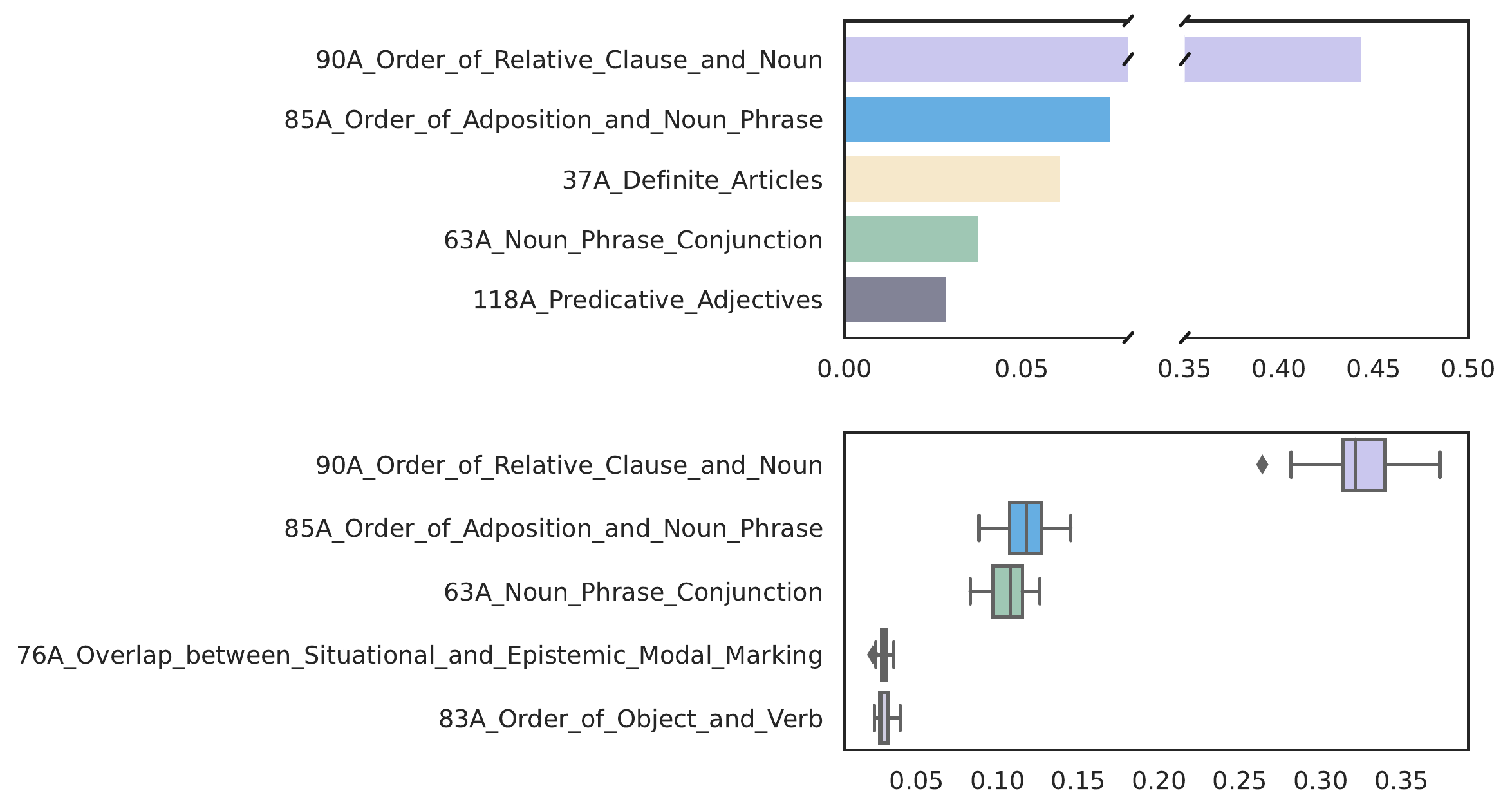}
  \caption{The five typological features having the biggest impact on the syntactic difference in mBERT. Above: Impurity-based importance. Below: Permutation importance.} \label{fig:feature-importance}
\end{figure}

Figure~\ref{fig:feature-importance} shows the five most important features\footnote{For importance of all features, see Appendix~\ref{sec:appendix-typo-feature-importance}.}. The dominant role of features belonging to the area of word order supports previous work emphasizing the importance of word order typology in characterizing the difference between languages \cite{ahmad_difficulties_2019, pires_how_2019, karthikeyan2019cross, dufter_identifying_2020}. 



\paragraph{Source language selection}

\begin{table}
\centering
\begin{tabular}{cc||cc}
\hline
\textbf{Method} & \textbf{NDCG(\%)} & \textbf{Method} & \textbf{NDCG(\%)} \\
\hline
\textsc{Ave}  & 66.1 & $d_{\textrm{geo}}$ & 52.6\\
$d_{\textrm{syn}}$ & 71.7 & $d_{\textrm{pho}}$ & 23.7\\
$d_{\textrm{gen}}$ & 61.6 & $d_{\textrm{inv}}$ & 59.4\\
$d_{\textrm{fea}}$ & 57.7 & \textbf{\textsc{Reg}} & \textbf{77.0}\\
\hline
\end{tabular}
\caption{Results of source language selection strategy evaluated by NDCG@3 (\%). \textsc{Reg} is our method.}
\label{tab:source-language-selection}
\end{table}

Our regressor effectively selects better source languages for zero-shot cross-lingual transfer of dependency parsing than baselines (Table~\ref{tab:source-language-selection}), which further verifies our findings and indicates that \textbf{morphosyntactic features are good indicators of transfer performance}.

\subsection{Discussion}
\label{sec:typology-vs-genealogy}

Previous work has tried to predict the cross-lingual transfer performance based on typological features \cite{lin_choosing_2019, pires_how_2019}, whereas a general metric of typological similarity may not be informative enough. \citet{dolicki_analysing_2021} conducted a finer-grained analysis, but the aim is to choose the best source language for a certain downstream task, not focusing on specific language pairs. 

We here show that the morphosyntactic features are predictive of the cross-linguistic syntactic difference learnt during pretraining and have a great potential to benefit the cross-lingual transfer. As our method is based on distributions and is not constrained at a language level, it can be extended to cross-domain and multi-source transfer scenarios, where data from different languages or domains can be treated as one dataset and the effects of linguistic properties may be reevaluated. Combined with finer-grained linguistic features, it is promising to provide more insight into the cross-lingual transfer.

\section{Related Work}
\label{sec:related-work}

\paragraph{Probing for linguistic knowledge}

Contextualized word embeddings have been found to be especially effective at the syntactic level \cite{linzen_syntactic_2021,baroni_proper_2021}. Through probing methods, prior work has shown that syntactic knowledge including syntactic tree depth, subject-verb agreement \cite{conneau_what_2018,jawahar_what_2019}, constituent labels, grammatical relations \cite{tenney_what_2018,liu_linguistic_2019} and dependency parse trees \cite{hewitt_structural_2019} can be largely derived from these embeddings. In the multilingual setting, mBERT has been found to encode morphosyntactic properties such as syntactic structure \cite{chi_finding_2020} and morphosyntactic alignment \cite{papadimitriou_deep_2021} in a similar way across languages. There has been work noting problems related to the probing method \cite{hewitt_designing_2019,pimentel_information-theoretic_2020,voita_information-theoretic_2020}, suggesting that the extra classifier can interfere with the analysis of the embedding space. We here derive representations of syntactic knowledge through a simple subtraction between embeddings and discard task-specific parameters through a measure of distance between their representations.

\paragraph{Linguistic diversity}

Difference in linguistic properties across languages has been associated with the hardness of transfer \cite{ponti_isomorphic_2018,lin_choosing_2019} and typological resources have been exploited to guide parameter and information sharing among languages \cite{naseem-etal-2012-selective,tackstrom-etal-2013-target,ammar-etal-2016-many} and data selection \cite{ponti_isomorphic_2018,lin_choosing_2019}. Previous work has demonstrated that the transfer performance is greatly affected by typological features such as word order both in a delexicalized setting before the emergence of large pretrained language models \cite{aufrant-etal-2016-zero} and in the context of multilingual language models \cite{pires_how_2019, karthikeyan2019cross, dufter_identifying_2020}. Moreover, much typological information is found encoded in mBERT representations \citep{choenni_what_2020} and blinding mBERT to it impedes successful cross-lingual transfer \citep{bjerva_does_2021}. On the other hand, \citet{singh_bert_2019} shows that the representation space of mBERT is partitioned in a way similar to genealogical relatedness. While most previous work investigates sentence-level or word-level representations and mixes various aspects of linguistic knowledge, we here focus on the cross-lingual syntactic transfer and extract representations in a targeted manner.

\section{Conclusion}
\label{sec:conclusion}

Languages vary profoundly at almost every level including lexicon, grammar and meaning. Pretrained multilingual encoders learn to encode them in a shared representation space simply via self-supervision, but it is unclear how they address the linguistic variation at different levels. This paper investigates the cross-lingual syntactic ability of mBERT. Through a measure of distance between distributions over its representations, we demonstrate that mBERT encodes universal grammatical relations in a way highly consistent with the cross-linguistic syntactic difference in terms of formal syntax. Such cross-linguistic syntactic knowledge plays a decisive role in the zero-shot cross-lingual transfer performance of dependency parsing. This evidence suggests that linguistic knowledge such as typological resources can be incorporated in improvement of cross-lingual transfer and thus help to better accommodate the rich linguistic diversity.

\section*{Limitations}

At the core of our method is a measure of divergence between distributions, which highly correlates with the zero-shot cross-lingual transfer performance. As it is challenging to choose an appropriate measure of divergence between joint distributions, we empirically compared several measures, and they yield similar results. We here employ the optimal transport distance between datasets \citep{alvarez-melis_geometric_2020} as it provides interpretable correspondence and characterize the geometry of the representation space. A detailed analysis of the best measure of divergence in the multilingual setting is left for future work.

Combined with representations of grammatical relations derived from mBERT, our method provides a quantitative evaluation of the cross-linguistic difference learnt by mBERT in terms of dependency grammar. It can be related with typological diversity and help to analyze the effects of various morphosyntactic properties. Future work can extend to finer-grained description of linguistic variation and other downstream tasks involving different aspects of language. By clarifying the source of cross-lingual transfer and understanding how linguistic diversity affects the model, significant improvements on efficient cross-lingual transfer can be expected.


\section*{Acknowledgements}

The authors would like to thank the anonymous reviewers for their helpful comments. This work was partially funded by National Natural Science Foundation of China (No. 62206057, 62076069, 61976056).

\bibliography{anthology,custom}
\bibliographystyle{acl_natbib}

\appendix

\section{Additional Materials for Measure of Syntactic Difference in mBERT}

\subsection{Representations for Grammatical Relations}
\label{sec:grammatical-relation-probe}

\paragraph{Grammatical relation probe}

For each language, we train a linear classifier via stochastic gradient descent\footnote{\url{https://scikit-learn.org/stable/modules/generated/sklearn.linear_model.SGDClassifier.html}} to identify the grammatical relation between a word pair $(w_{\textrm{\footnotesize head}}, w_{\textrm{\footnotesize dep}})$ given the input representation $\mathbf{d}_{(\textrm{\footnotesize head}, \textrm{\footnotesize dep})}^{\ell}$.

Table~\ref{tab:probe-comparison-baseline} shows that the layer 7 of mBERT significantly outperforms the baselines in representing grammatical relations ($W = 0.0$ and $p = 1.19 \times 10^{-7}$)\footnote{As \textsc{mBERTrand} performs similar across different layers, we take the 7th layer of it for comparison in the following experiments.}.

\begin{table*}
\centering
\begin{tabular}{lcccccccccccc}
\hline
Language & ar & bg & de & el & en & es & et & fa & fi & fr & he & hi \\
\hline
Layer 7 & 83.3 & 89.3 & 87.0 & 92.0 & 88.0 & 88.9 & 77.3 & 88.0 & 79.3 & 90.1 & 85.8 & 86.0 \\
\textsc{mBERT0} & 60.1 & 60.7 & 68.2 & 61.1 & 69.2 & 66.4 & 47.8 & 62.7 & 49.9 & 67.8 & 58.2 & 60.2 \\
\textsc{mBERTrand} & 58.8 & 61.6 & 69.0 & 62.5 & 70.2 & 67.4 & 47.7 & 61.9 & 50.5 & 68.3 & 59.1 & 60.8 \\
\hline
\hline
Language & it & ja & ko & lv & nl & pl & pt & ro & ru & tr & vi & zh \\
\hline
Layer 7 & 88.7 & 86.5 & 77.8 & 79.8 & 87.1 & 86.4 & 92.5 & 86.4 & 89.2 & 73.7 & 70.8 & 84.3 \\
\textsc{mBERT0} & 65.6 & 61.2 & 45.4 & 50.6 & 62.2 & 54.8 & 68.4 & 56.3 & 60.9 & 50.1 & 58.6 & 56.3 \\
\textsc{mBERTrand} & 66.0 & 61.6 & 46.2 & 51.1 & 63.8 & 57.0 & 69.1 & 58.2 & 61.9 & 51.2 & 59.6 & 57.9 \\
\hline
\end{tabular}
\caption{Comparison of the 7th layer of mBERT and the two baselines. We take the best layer of \textsc{mBERTrand} for comparison.}
\label{tab:probe-comparison-baseline}
\end{table*}

\paragraph{Visualization of the representation space}

We combine t-SNE\footnote{\url{https://scikit-learn.org/stable/modules/generated/sklearn.manifold.TSNE.html}} \cite{JMLR:v9:vandermaaten08a} with PCA to visualize the representations in two dimensions\footnote{As t-SNE can be slow for high-dimensional data, the representations are first projected to 37 dimensions via PCA and then visualized through t-SNE.}. As to t-SNE, the perplexity is set to 30 and the maximum number of iteration is set to 1000.

\subsection{Evaluation of Syntactic Difference in mBERT}

\paragraph{Distance between Distributions}
\label{sec:appendix-otdd}

The method of optimal transport dataset distance (OTDD) \cite{alvarez-melis_geometric_2020} relies on optimal transport and defines the metric space as $\mathcal{Z}= \mathcal{X} \times \mathcal{Y}$, where $\mathcal{X}$ is the feature space and $\mathcal{Y}$ is the label set. The metric on $\mathcal{Z}$ is defined as $$d_{\mathcal{Z}} \left(\mathbf{z}, \mathbf{z}^{\prime} \right) = \left(d_{\mathcal{X}}\left(\mathbf{x}, \mathbf{x}^{\prime} \right)^{p} + d_{\mathcal{Y}}\left(\mathbf{y}, \mathbf{y}^{\prime} \right)^{p} \right)^{1/p}$$ where $\mathbf{z} = (\mathbf{x}, \mathbf{y})$ is a feature-label pair and $p \ge 1$. Euclidean distance is employed for metric $d_{\mathcal{X}}$ on the feature space $\mathcal{X}$. For $d_{\mathcal{Y}}$, labels are regarded as distributions over $\mathcal{X}$ where samples with label $\mathbf{y}$ are drawn. $d_{\mathcal{Y}}$ is measured through $p$-Wasserstein distance between distributions of labels. The distance between dataset $\mathcal{D}$ and $\mathcal{D}^{\prime}$ is calculated as $$d_{\mathrm{OT}}\left(\mathcal{D}, \mathcal{D}^{\prime}\right)=\min _{\pi \in \Pi(\mathcal{D}, \mathcal{D}^{\prime})} \int_{\mathcal{Z} \times \mathcal{Z}} d_{\mathcal{Z}}\left(\mathbf{z}, \mathbf{z}^{\prime}\right) \pi(\mathbf{z}, \mathbf{z}^{\prime})$$ where $\pi$ is the coupling matrix. For more details, see \citet{alvarez-melis_geometric_2020}.

\subsection{Validation of Syntactic Difference in mBERT}

\begin{figure}[t]
\centering
  \includegraphics[width=7.5cm]{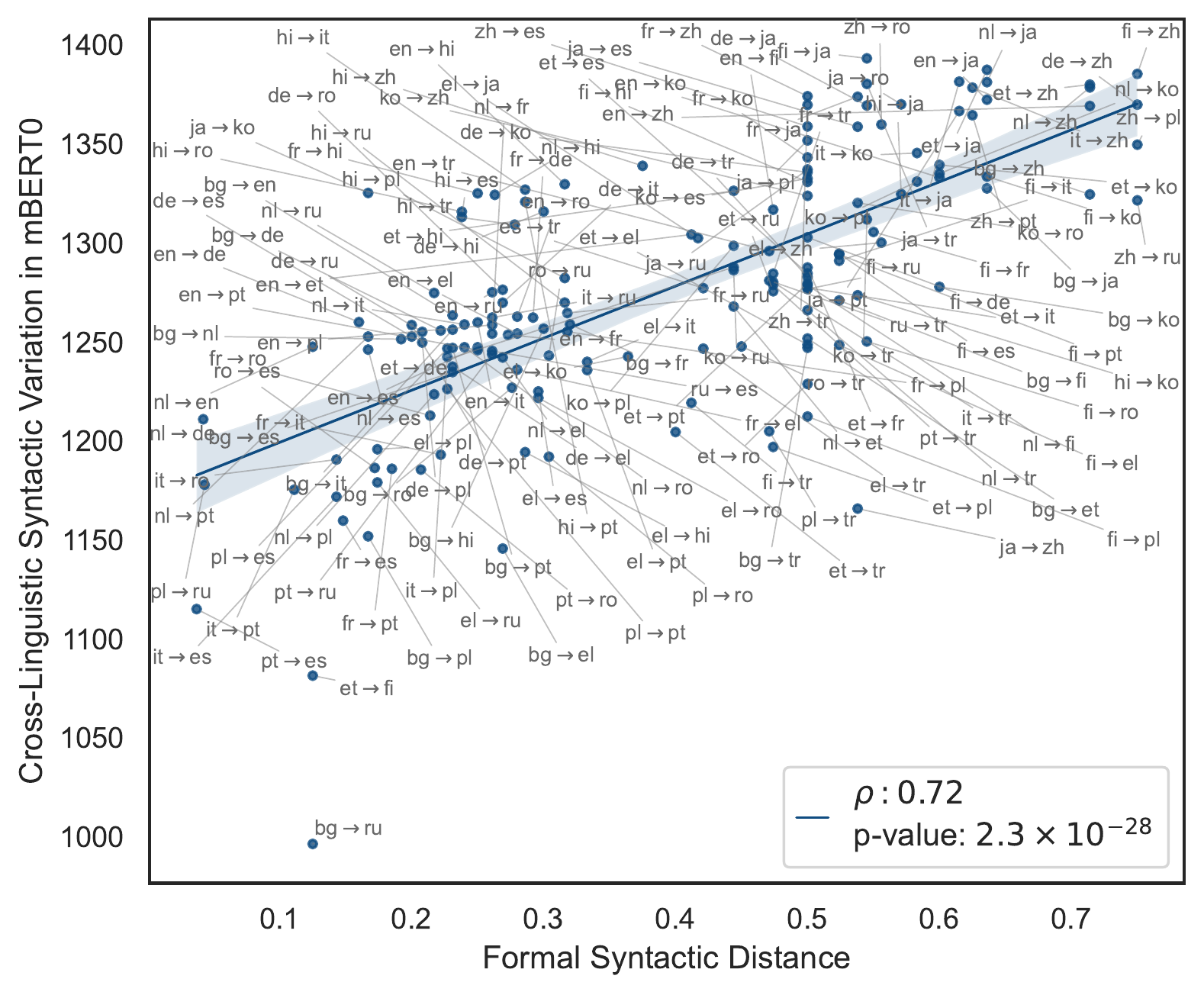}
  \caption{Comparison of formal syntactic distance and cross-linguistic syntactic difference derived from \textsc{mBERT0}.} \label{fig:baseline-formal-otdd-mbert0}
\end{figure}

\begin{figure}[t]
\centering
  \includegraphics[width=7.5cm]{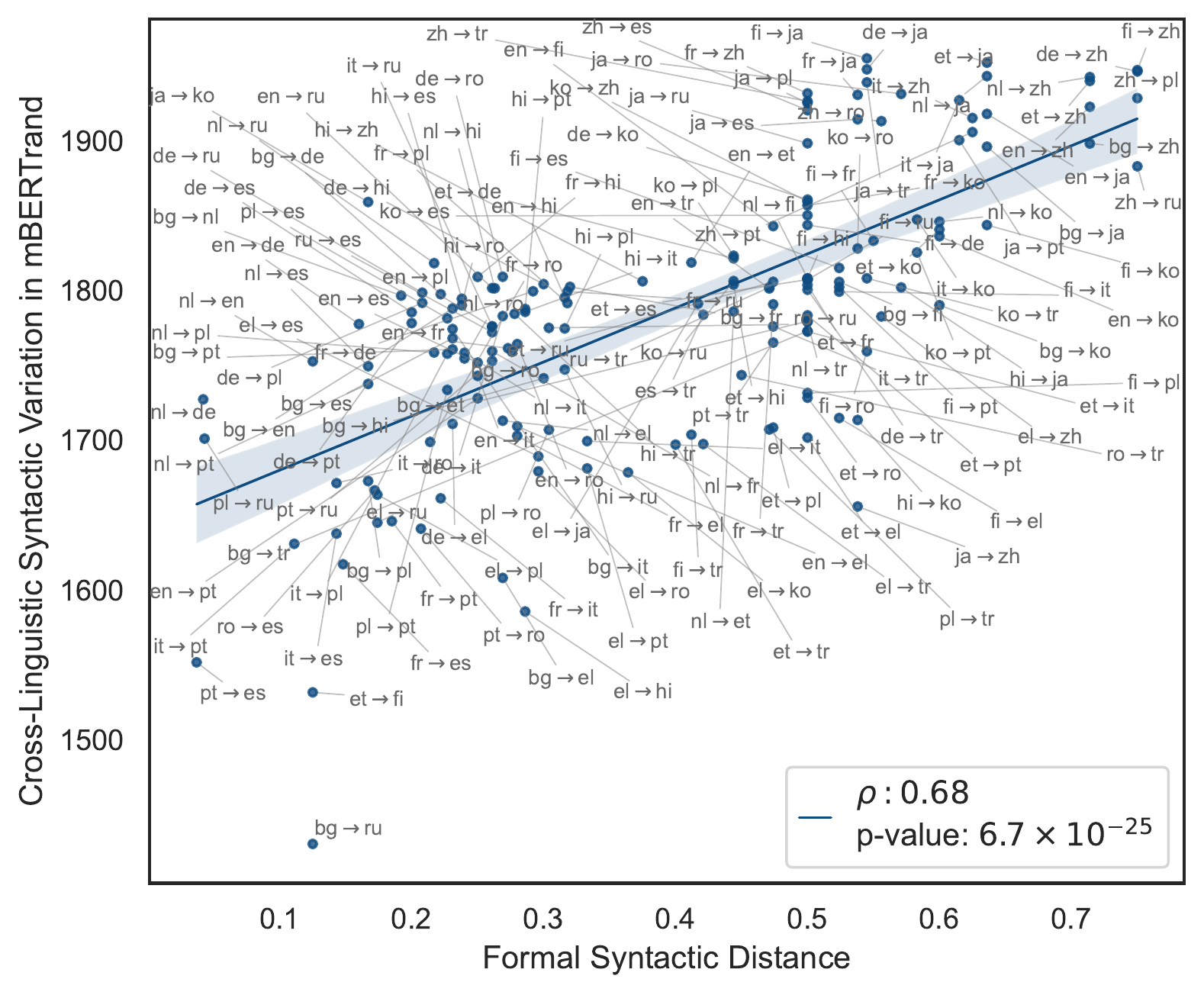}
  \caption{Comparison of formal syntactic distance and cross-linguistic syntactic difference derived from \textsc{mBERTrand}.} \label{fig:baseline-formal-otdd-mbertrand}
\end{figure}

Figure~\ref{fig:baseline-formal-otdd-mbert0} and Figure~\ref{fig:baseline-formal-otdd-mbertrand} show the comparison of syntactic difference derived from two baselines and the formal syntactic distance. The lower Spearman's $\rho$ indicates that the similarities and differences between languages are not well captured by these baselines.

\section{Additional Materials for Mechanism behind Cross-Lingual Transfer}

\subsection{Comparison with Baselines}

\begin{figure}[t!]
\centering
  \includegraphics[width=7.5cm]{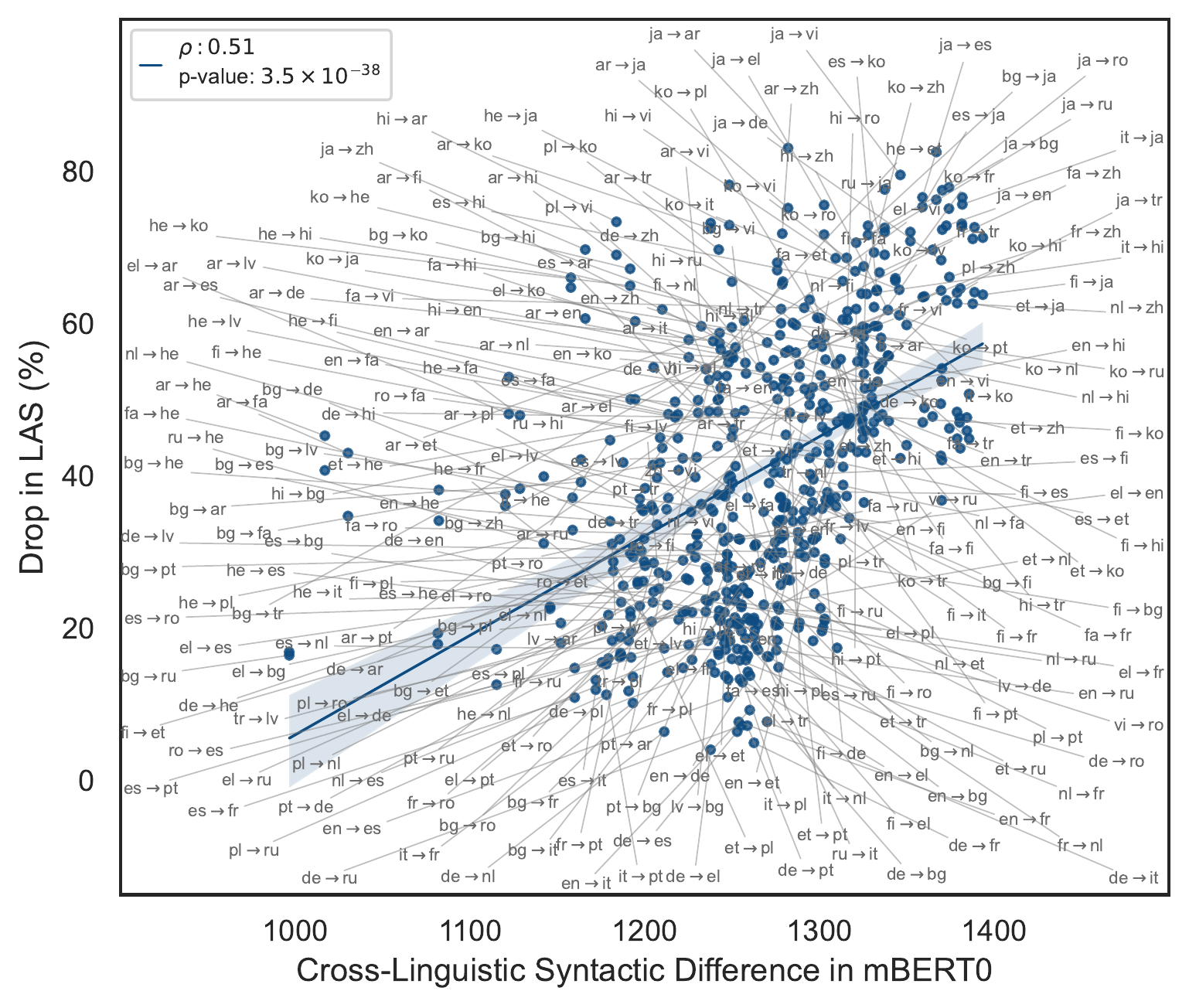}
  \caption{Comparison of the cross-linguistic syntactic difference in \textsc{mBERT0} and drop in zero-shot cross-lingual transfer performance (LAS).} \label{fig:baseline-otdd-mbert0-las-all-langs}
\end{figure}

\begin{figure}[t!]
\centering
  \includegraphics[width=7.5cm]{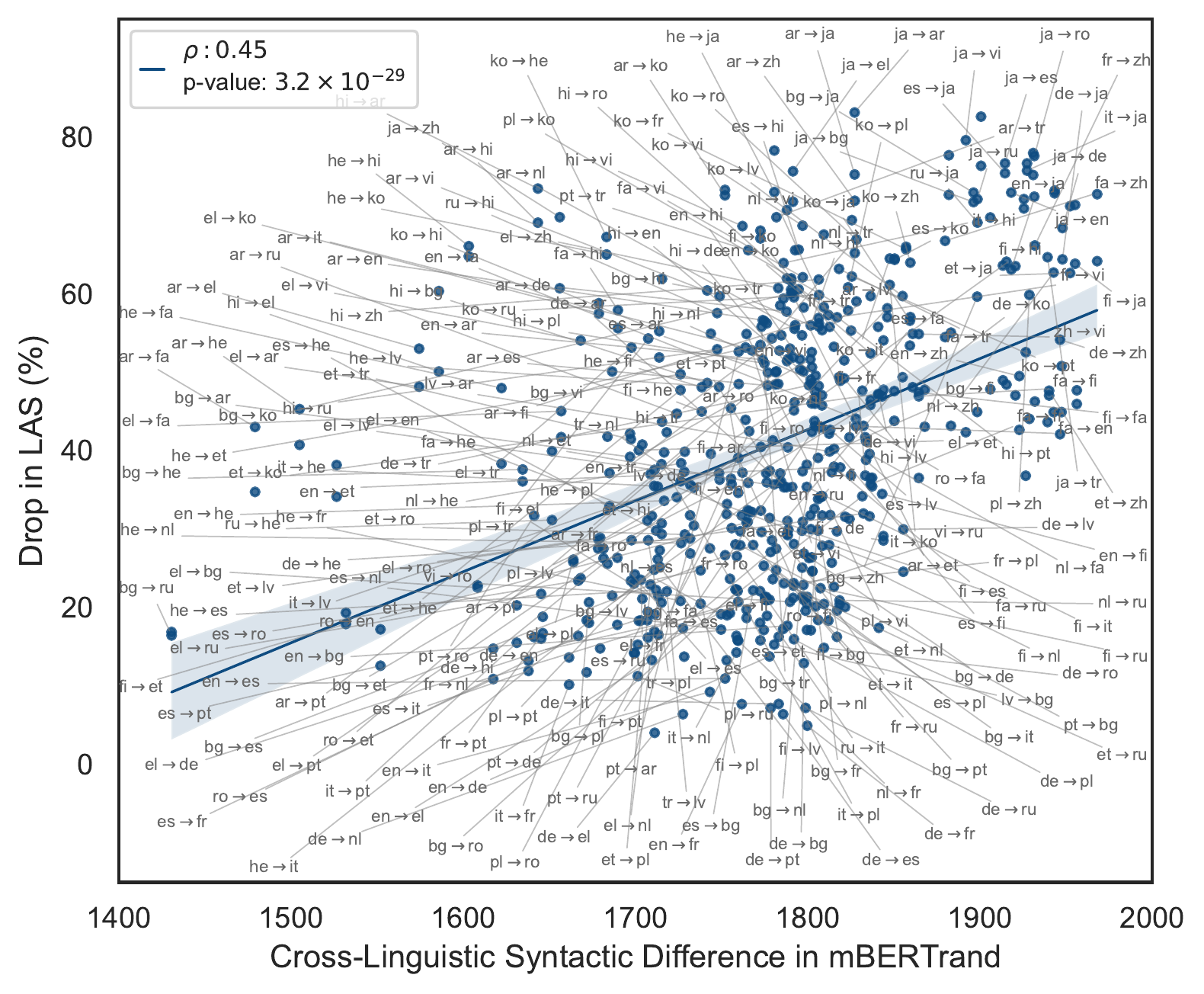}
  \caption{Comparison of the cross-linguistic syntactic difference in \textsc{mBERTrand} and drop in zero-shot cross-lingual transfer performance (LAS).} \label{fig:baseline-otdd-mbertrand-las-all-langs}
\end{figure}

Figure~\ref{fig:baseline-otdd-mbert0-las-all-langs} and Figure~\ref{fig:baseline-otdd-mbertrand-las-all-langs} are comparisons of the syntactic difference induced by two baselines and the performance drop in zero-shot cross-lingual transfer performance of dependency parsing (Section~\ref{sec:mechanism-result}).

\section{Additional Materials for Impact of Linguistic Diversity}

\subsection{Typological Features}
\label{sec:appendix-typo}

Table~\ref{tab:wals-features} shows the morphosyntactic features we employ in Section~\ref{sec: typology}. We delete Feature 95A: \emph{Relationship between the Order of Object and Verb and the Order of Adposition and Noun Phrase}, 96A:  \emph{Relationship between the Order of Object and Verb and the Order of Relative Clause and Noun} and 97A: \emph{Relationship between the Order of Object and Verb and the Order of Adjective and Noun} as they can be inferred from other features in the area of word order. 

\subsection{Importance of Typological Features}
\label{sec:appendix-typo-feature-importance}

\begin{figure*}[t!]
\centering
  \includegraphics[width=16.4cm]{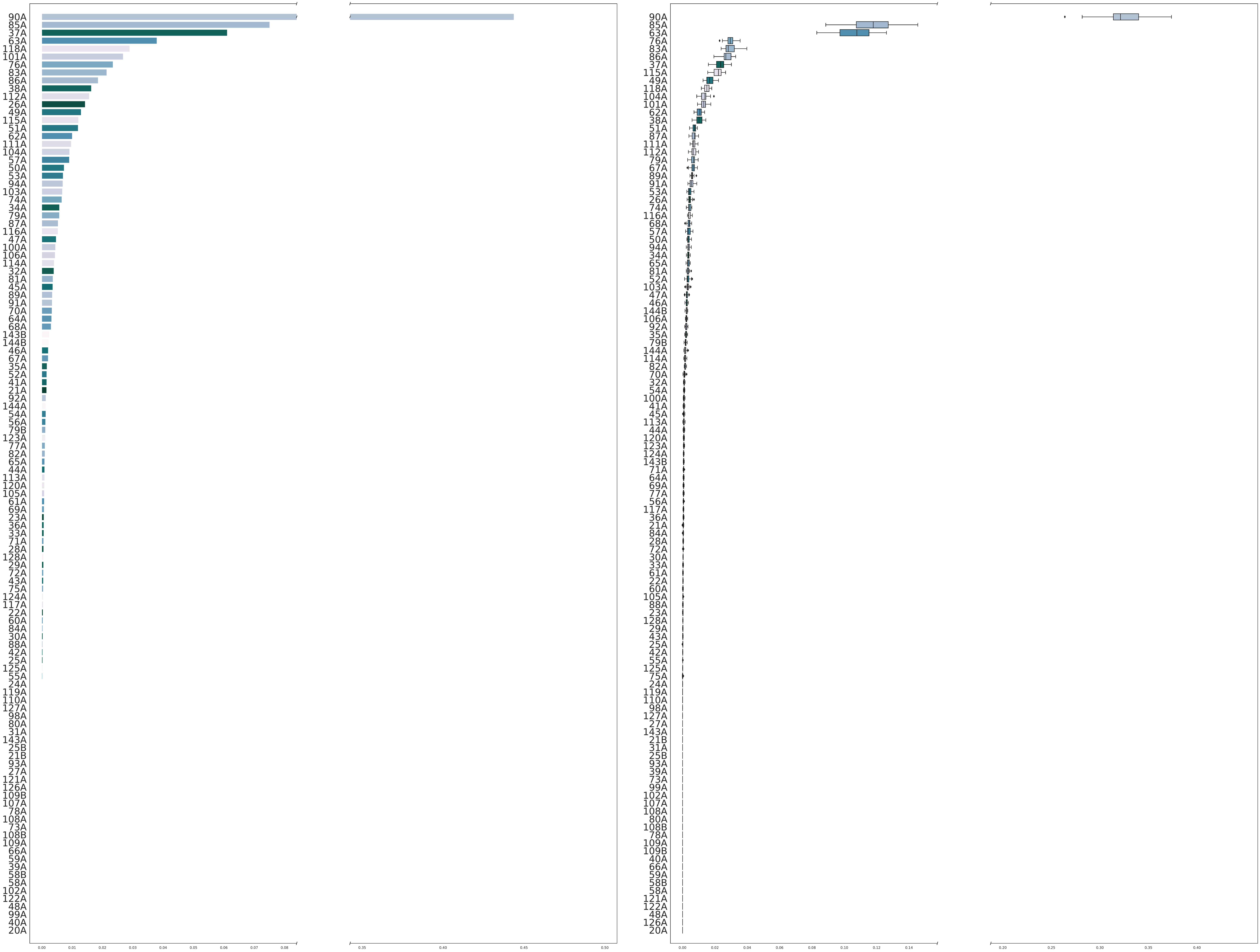}
  \caption{Rank of importance of all the morphosyntactic features we use.} \label{fig:all-feature-importance}
\end{figure*}

The feature importance of all the morphosyntactic features we use is shown in Figure~\ref{fig:all-feature-importance}.

\section{Implementation Details}

\paragraph{Multilingual BERT}
We use the pretrained \emph{bert-base-multilingual-cased} model\footnote{\url{https://huggingface.co/bert-base-multilingual-cased}} for all our experiments.

\paragraph{Grammatical relation probe}

We train the linear classifier via stochastic gradient descent\footnote{\url{https://scikit-learn.org/stable/modules/generated/sklearn.linear_model.SGDClassifier.html}} to classify the grammatical relations between a head-dependent word pair. We use logistic regression, set the max number of iteration to 10000 and allow for early stopping. We report the $95\%$ confidence interval computed based on different regularization strengths (1.e-09, 1.e-08, 1.e-07, 1.e-06, 1.e-05, 1.e-04, 1.e-03, and 1.e-02) in Figure~\ref{fig:probe-acc}.

\paragraph{Measure of the syntactic difference in mBERT}

We use the public source code of \citet{alvarez-melis_geometric_2020}\footnote{\url{https://github.com/microsoft/otdd}} to compute the syntactic difference in mBERT. The $p$-Wasserstein distance ($p=2$) is computed based on Sinkhorn algorithm \cite{cuturi_sinkhorn_2013} and the entropy regularization strength is set to 1e-1. 

\paragraph{Dependency parsing}

We follow the setup of \citet{wu_beto_2019}, which replaces the LSTM encoder in \citet{dozat_deep_2017} with mBERT. For each language, we train the model with ten epochs and validate it at the end of each epoch. We choose the model performing the best (i.e., achieving the highest LAS) on the development set. We use the Adam optimizer with $\beta_{1} = 0.9$, $\beta_{2} = 0.99$, $\textrm{eps} = 1 \times 10^{-8}$, and a learning rate of 5e-5. The batch size is 16 and the max sequence length is 128.

\paragraph{Gradient boosting regressor}

We use a gradient boosting regressor\footnote{\url{https://scikit-learn.org/stable/modules/generated/sklearn.ensemble.GradientBoostingRegressor.html}} with 100 estimators and each has a maximum depth of 3. We use the squared error for regression with the default learning rate of 1e-1.

\section{Data for Experiments}

\subsection{Universal Dependencies Treebanks}
\label{sec:data}

Table~\ref{tab:ud-treebanks} shows the languages and UD treebanks (version 2.8)\footnote{\url{https://lindat.mff.cuni.cz/repository/xmlui/bitstream/handle/11234/1-3687/ud-treebanks-v2.8.tgz?sequence=1&isAllowed=y}} we use. We follow the split of training, development and test set in UD.


\begin{table*}
\centering
\begin{tabular}{lcllrr}
\hline
\textbf{Language} & \textbf{Abbr.} & \textbf{Language Family} & \textbf{UD Treebanks} & \textbf{\#Sentences} & \textbf{\#Tokens}\\
\hline
$\textrm{Arabic}^{\dagger}$ & $\textrm{ar}^{\dagger}$ & Afro-Asiatic.Semitic & $\textrm{Arabic-PADT}^{\dagger}$ & 7,664 & 282,384\\
Bulgarian & bg & IE.Balto-Slavik & Bulgarian-BTB & 11,138 & 156,149\\
$\textrm{Catalan}^{\ast}$ & $\textrm{ca}^{\ast}$ & IE.Romance & $\textrm{Catalan-AnCora}^{\ast}$ & 166,678 & 530,767 \\
$\textrm{Czech}^{\ast}$ & $\textrm{cs}^{\ast}$ & IE.Balto-Slavic & $\textrm{Czech-PDT}^{\ast}$ & 87,913 & 1,503,732 \\
$\textrm{German}$ & $\textrm{de}$ & IE.Germanic & $\textrm{German-GSD}$ & 15,590 & 287,740 \\
Greek & el & IE.Greek & Greek-GDT & 2,521 & 61,773 \\
English & en & IE.Germanic & English-EWT & 16,621 & 251,494 \\
Spanish & es & IE.Romance & Spanish-GSD & 16,013 & 423,346 \\
Estonian & et & Uralic & Estonian-EDT & 30,972 & 437,767 \\
$\textrm{Persian (Farsi)}^{\dagger}$ & $\textrm{fa}^{\dagger}$ & IE.Indo-Iranian & $\textrm{Persian-PerDT}^{\dagger}$ & 29,107 & 494,163 \\
Finnish & fi & Uralic & Finnish-TDT & 151,136 & 201,950 \\
French & fr & IE.Romance & French-GSD & 16,341 & 389,224 \\
$\textrm{Hebrew}^{\dagger}$ & $\textrm{he}^{\dagger}$ & Afro-Asiatic.Semitic & $\textrm{Hebrew-HTB}^{\dagger}$ & 6,216 & 115,529 \\
Hindi & hi & IE.Indo-Iranian & Hindi-HDTB & 16,647 & 351,704 \\
$\textrm{Hungarian}^{\ast}$ & $\textrm{hu}^{\ast}$ & Uralic & $\textrm{Hungarian-Szeged}^{\ast}$ & 1,800 & 42,032 \\
Italian & it & IE.Romance & Italian-VIT & 10,087 & 259,479 \\
Japanese & ja & Japonic & Japanese-GSD & 8,100 & 193,654 \\
Korean & ko & Koreanic & Korean-Kaist & 27,363 & 350,090 \\
$\textrm{Latvian}^{\dagger}$ & $\textrm{lv}^{\dagger}$ & IE.Balto-Slavic & $\textrm{Latvian-LVTB}^{\dagger}$ & 15,351 & 252,334 \\
Dutch & nl & IE.Germanic & Dutch-Alpino & 13,603 & 208,613 \\
Polish & pl & IE.Balto-Slavic & Polish-PDB & 22,152 & 347,377 \\
Portuguese & pt & IE.Romance & Portuguese-GSD & 12,078 & 297,938 \\
Romanian & ro & IE.Romance & Romanian-RRT & 9,524 & 218,511 \\
Russian & ru & IE.Balto-Slavic & Russian-GSD & 5,030 & 98,000 \\
$\textrm{Tamil}^{\ast}$ & $\textrm{ta}^{\ast}$ & Dravidian & $\textrm{Tamil-TTB}^{\ast}$ & 600 & 8,635 \\
Turkish & tr & Turkic & Turkish-BOUN & 9,761 & 121,214 \\
$\textrm{Urdu}^{\ast}$ & $\textrm{ur}^{\ast}$ & IE.Indo-Iranian & $\textrm{Urdu-UDTB}^{\ast}$ & 5,130 & 138,077 \\
$\textrm{Vietnamese}^{\dagger}$ & $\textrm{vi}^{\dagger}$ & Austroasiatic.Vietic & $\textrm{Vietnamese-VTB}^{\dagger}$ & 3,000 & 43,754 \\
Chinese (Mandarin) & zh & Sino-Tibetan.Sinitic & Chinese-GSDSimp & 4,997 & 123,291 \\
\hline
\end{tabular}
\caption{\label{tab:ud-treebanks}
Languages and UD Treebanks we use. Languages marked with a dagger ($\dagger$) aren't involved in the comparison with formal syntactic distance due to lack of corresponding data in \citet{ceolin_formal_2020}. Languages used for test of the strategy for source language selection in Section~\ref{sec: typology} is marked with an asterisk ($\ast$). The phylogenetic information is obtained from Glottolog \cite{harald_hammarstrom_2021_5772642}. IE stands for the Indo-European family.}
\end{table*}

\clearpage
\onecolumn
\begin{longtable}[c]{lp{11cm}cc}
\caption{Features in WALS used in our work. As WALS entries can be sparse, we provide in the column \textbf{\#Languages} information about how many languages involved in the experiment (Section~\ref{sec: typology}) have a valid entry for the feature. The \textbf{left} side of "/" indicates the number of languages for which the feature is not missing among the languages involved in the training procedure of the gradient boosting regressor, including Arabic, Bulgarian, German, Greek, English, Spanish, Estonian, Persian, Finnish, French, Hebrew, Hindi, Italian, Japanese, Korean, Latvian, Dutch, Polish, Portuguese, Romanian, Russian, Turkish and Chinese. For the \textbf{right} side, the five languages used to test the strategy for source language selection is involved, including Czech, Catalan, Hungarian, Tamil and Urdu.}

\label{tab:wals-features}\\
\toprule
\textbf{ID} & \textbf{Name} & \textbf{\#Languages}\\
\midrule
\endfirsthead 
\toprule
\textbf{ID} & \textbf{Name} & \textbf{\#Languages}\\
\midrule
\endhead
20A & Fusion of Selected Inflectional Formatives & 15 / 16\\
21A & Exponence of Selected Inflectional Formatives & 15 / 16\\
21B & Exponence of Tense-Aspect-Mood Inflection & 15 / 16\\
22A & Inflectional Synthesis of the Verb & 15 / 16 \\
23A & Locus of Marking in the Clause & 15 / 16\\
24A	& Locus of Marking in Possessive Noun Phrases & 15 / 16\\
25A	& Locus of Marking: Whole-language Typology	& 15 / 16\\
25B & Zero Marking of A and P Arguments	& 15 / 16\\
26A	& Prefixing vs. Suffixing in Inflectional Morphology & 23 / 27\\
27A	& Reduplication	& 17 / 20\\
28A	& Case Syncretism & 16 / 17\\
29A	& Syncretism in Verbal Person/Number Marking & 16 / 17\\
30A	& Number of Genders	& 14 / 16\\
31A	& Sex-based and Non-sex-based Gender Systems & 14 / 16\\
32A	& Systems of Gender Assignment & 14 / 16\\
33A	& Coding of Nominal Plurality & 23 / 27\\
34A	& Occurrence of Nominal Plurality & 18 / 20\\
35A	& Plurality in Independent Personal Pronouns & 16 / 18\\
36A	& The Associative Plural & 21 / 22\\
37A	& Definite Articles	& 22 / 25\\
38A	& Indefinite Articles & 20 / 24\\
39A & Inclusive/Exclusive Distinction in Independent Pronouns & 16 / 17\\
40A & Inclusive/Exclusive Distinction in Verbal Inflection & 16 / 17\\
41A & Distance Contrasts in Demonstratives & 16 / 20\\	
42A & Pronominal and Adnominal Demonstratives & 16 / 20\\
43A	& Third Person Pronouns and Demonstratives & 14 / 15\\
44A	& Gender Distinctions in Independent Personal Pronouns & 19 / 20\\
45A	& Politeness Distinctions in Pronouns & 21 / 24\\
46A & Indefinite Pronouns & 23 / 25\\
47A & Intensifiers and Reflexive Pronouns & 22 / 25\\
48A & Person Marking on Adpositions	& 19 / 20\\
49A	& Number of Cases & 21 / 24\\	
50A & Asymmetrical Case-Marking	& 21 / 24\\
51A & Position of Case Affixes & 23 / 27\\
52A & Comitatives and Instrumentals & 20 / 24\\
53A & Ordinal Numerals & 23 / 27\\
54A & Distributive Numerals & 20 / 23\\
55A	& Numeral Classifiers & 15 / 16\\
56A	& Conjunctions and Universal Quantifiers & 12 / 14\\	
57A & Position of Pronominal Possessive Affixes	& 18 / 20\\
58A & Obligatory Possessive Inflection & 15 / 16\\
58B	& Number of Possessive Nouns & 15 / 16\\
59A	& Possessive Classification & 15 / 16\\
60A	& Genitives, Adjectives and Relative Clauses & 11 / 12\\
61A	& Adjectives without Nouns & 13 / 14\\
62A & Action Nominal Constructions & 19 / 21\\
63A	& Noun Phrase Conjunction & 20 / 22\\
64A	& Nominal and Verbal Conjunction & 17 / 19\\
65A & Perfective/Imperfective Aspect & 19 / 21\\
66A	& The Past Tense & 19 / 21\\
67A & The Future Tense & 19 / 21\\
68A	& The Perfect & 19 / 21\\
69A	& Position of Tense-Aspect Affixes & 23 / 27\\
70A	& The Morphological Imperative & 23 / 27\\
71A	& The Prohibitive & 23 / 27\\
72A	& Imperative-Hortative Systems & 23 / 26\\
73A	& The Optative & 18 / 21\\
74A & Situational Possibility & 21 / 24\\
75A	& Epistemic Possibility & 21 / 24\\
76A & Overlap between Situational and Epistemic Modal Marking & 21 / 24\\
77A	& Semantic Distinctions of Evidentiality & 20 / 22\\
78A	& Coding of Evidentiality & 20 / 22\\
79A	& Suppletion According to Tense and Aspect& 19 / 21\\
79B	& Suppletion in Imperatives and Hortatives & 19 / 21\\
80A	& Verbal Number and Suppletion & 19 / 21\\
81A & Order of Subject, Object and Verb & 23 / 28\\
82A & Order of Subject and Verb & 23 / 28\\
83A & Order of Object and Verb & 23 / 28\\
84A & Order of Object, Oblique, and Verb & 12 / 13\\	
85A & Order of Adposition and Noun Phrase & 23 / 28\\	
86A & Order of Genitive and Noun & 23 / 28\\
87A & Order of Adjective and Noun & 23 / 28\\
88A & Order of Demonstrative and Noun & 23 / 28\\
89A & Order of Numeral and Noun & 22 / 27\\
90A & Order of Relative Clause and Noun & 23 / 28\\
91A & Order of Degree Word and Adjective & 22 / 25\\
92A & Position of Polar Question Particles & 23 / 27\\
93A & Position of Interrogative Phrases in Content Questions & 20 / 24\\
94A & Order of Adverbial Subordinator and Clause & 21 / 25\\	
98A & Alignment of Case Marking of Full Noun Phrases & 16 / 17\\
99A & Alignment of Case Marking of Pronouns & 16 / 17\\
100A & Alignment of Verbal Person Marking & 19 / 20\\
101A & Expression of Pronominal Subjects & 21 / 24\\
102A & Verbal Person Marking & 19 / 20\\
103A & Third Person Zero of Verbal Person Marking & 19 / 20\\
104A & Order of Person Markers on the Verb & 19 / 20\\	
105A & Ditransitive Constructions: The Verb 'Give' & 17 / 19\\
106A & Reciprocal Constructions & 17 / 18\\
107A & Passive Constructions & 19 / 20\\
108A & Antipassive Constructions & 16 / 18\\
108B & Productivity of the Antipassive Construction & 16 / 18\\
109A & Applicative Constructions & 16 / 18\\
109B & Other Roles of Applied Objects & 16 / 18\\
110A & Periphrastic Causative Constructions & 13 / 15\\
111A & Nonperiphrastic Causative Constructions & 16 / 18\\
112A & Negative Morphemes & 23 / 27\\
113A & Symmetric and Asymmetric Standard Negation & 17 / 18\\
114A & Subtypes of Asymmetric Standard Negation & 17 / 18\\
115A & Negative Indefinite Pronouns and Predicate Negation & 22 / 25\\
116A & Polar Questions & 23 / 28\\
117A & Predicative Possession & 17 / 20\\
118A & Predicative Adjectives & 20 / 23\\
119A & Nominal and Locational Predication & 20 / 23\\
120A & Zero Copula for Predicate Nominals & 20 / 23\\
121A & Comparative Constructions & 15 / 17\\
122A & Relativization on Subjects & 18 / 19\\
123A & Relativization on Obliques & 18 / 19\\
124A & 'Want' Complement Subjects & 18 / 19\\
125A & Purpose Clauses & 15 / 17\\
126A & 'When' Clauses & 16 / 18\\
127A & Reason Clauses & 16 / 18\\
128A & Utterance Complement Clauses & 15 / 17\\
143A & Order of Negative Morpheme and Verb & 23 / 28\\
143B & Obligatory Double Negation & 23 / 28\\
144A & Position of Negative Word With Respect to Subject, Object, and Verb & 23 / 28\\
144B & Position of negative words relative to beginning and end of clause and with respect to adjacency to verb & 23 / 28\\
\bottomrule
\end{longtable}
\clearpage
\twocolumn

\end{document}